\documentclass[9pt,a4paper,twocolumn,twoside]{rho-class/rho}
\usepackage[english]{babel}

\usepackage{amsfonts}
\usepackage{amsmath}
\usepackage{amssymb}
\usepackage{array}
\usepackage{cuted}
\usepackage{placeins}
\usepackage{tabularx}
\usepackage{tikz}
\usepackage{wrapfig}

\usepackage{capt-of}

\newcommand{\PaperTableStyle}{%
  \small\rmfamily
  \setlength{\tabcolsep}{5pt}%
  \renewcommand{\arraystretch}{1.10}%
}
\newcommand{\CompactTableStyle}{%
  \small\rmfamily
  \setlength{\tabcolsep}{4pt}%
  \renewcommand{\arraystretch}{1.08}%
}
\AtBeginEnvironment{tabular}{\PaperTableStyle}
\AtBeginEnvironment{tabular*}{\PaperTableStyle}
\AtBeginEnvironment{tabularx}{\PaperTableStyle}
\makeatletter
\g@addto@macro\normalsize{%
  \setlength{\abovedisplayskip}{12pt plus 3pt minus 2pt}%
  \setlength{\belowdisplayskip}{12pt plus 3pt minus 2pt}%
  \setlength{\abovedisplayshortskip}{9pt plus 2pt minus 2pt}%
  \setlength{\belowdisplayshortskip}{10pt plus 2pt minus 2pt}%
  \setlength{\jot}{0.45em}%
}
\makeatother

\newcolumntype{Y}{>{\raggedright\arraybackslash}X}

\definecolor{theme}{HTML}{50B4A8}
\newcommand{\cI}{\cellcolor{theme!20}}
\newcommand{\cII}{\cellcolor{theme!14}}
\newcommand{\cIII}{\cellcolor{theme!9}}
\newcommand{\cIV}{\cellcolor{theme!6}}
\newcommand{\cV}{\cellcolor{theme!3}}
\newcommand{\dajilogomark}{%
  \tikz[baseline=-0.72ex,x=0.022pt,y=-0.022pt]{%
    \draw[line width=0.32pt,line cap=round,line join=round]
      (210,365)
      .. controls (170,345) and (150,300) .. (166,252)
      .. controls (178,215) and (200,185) .. (232,156)
      .. controls (262,129) and (302,120) .. (338,131)
      .. controls (375,143) and (403,171) .. (415,208)
      .. controls (428,247) and (423,292) .. (404,327)
      .. controls (382,368) and (341,391) .. (286,392)
      .. controls (253,392) and (228,385) .. (210,365);
    \draw[line width=0.32pt,line cap=round,line join=round]
      (252,136)
      .. controls (245,114) and (261,97) .. (282,98)
      .. controls (292,98) and (301,103) .. (308,113)
      .. controls (313,96) and (327,86) .. (344,88)
      .. controls (366,91) and (378,111) .. (372,132);
    \draw[line width=0.32pt,line cap=round,line join=round]
      (315,246)
      .. controls (325,235) and (344,235) .. (354,246)
      .. controls (346,262) and (333,270) .. (321,270)
      .. controls (313,263) and (309,255) .. (315,246);
    \fill (294,228) circle (5);
    \fill (372,228) circle (5);
    \draw[line width=0.32pt,line cap=round,line join=round]
      (210,208)
      .. controls (172,188) and (136,188) .. (118,206)
      .. controls (104,220) and (106,239) .. (124,247)
      .. controls (111,255) and (112,273) .. (129,282)
      .. controls (149,293) and (180,290) .. (202,278);
    \draw[line width=0.32pt,line cap=round,line join=round]
      (410,262)
      .. controls (432,256) and (452,265) .. (458,283)
      .. controls (462,296) and (458,309) .. (446,316)
      .. controls (457,323) and (458,337) .. (449,346)
      .. controls (437,358) and (417,357) .. (401,344);
    \draw[line width=0.32pt,line cap=round,line join=round] (238,387) -- (212,414);
    \draw[line width=0.32pt,line cap=round,line join=round] (212,414) -- (198,406);
    \draw[line width=0.32pt,line cap=round,line join=round] (212,414) -- (202,428);
    \draw[line width=0.32pt,line cap=round,line join=round] (212,414) -- (190,417);
    \draw[line width=0.32pt,line cap=round,line join=round] (332,389) -- (360,414);
    \draw[line width=0.32pt,line cap=round,line join=round] (360,414) -- (377,406);
    \draw[line width=0.32pt,line cap=round,line join=round] (360,414) -- (370,428);
    \draw[line width=0.32pt,line cap=round,line join=round] (360,414) -- (383,417);
  }%
}

\doctype{DAJI\hspace{0.45em}\raisebox{1.05ex}{\dajilogomark}}
\title{Before the Body Moves: Learning Anticipatory Joint Intent for Language-Conditioned Humanoid Control}

\author[1,2]{Haozhe Jia\textsuperscript{\ensuremath{\dagger}}}
\author[1]{Honglei Jin\textsuperscript{\ensuremath{\dagger}}}
\author[3]{Yuan Zhang\textsuperscript{\ensuremath{\dagger}}}
\author[1]{Youcheng Fan\textsuperscript{\ensuremath{\dagger}}}
\author[1]{Shaofeng Liang}
\author[5,6]{Lei Wang}
\author[3]{Shuxu Jin}
\author[1]{Kuimou Yu}
\author[3]{Zinuo Zhang}
\author[2]{Jianfei Song}
\author[1,2]{Wenshuo Chen\textsuperscript{*}}
\author[1,4]{Yutao Yue\textsuperscript{*}}

\affil[1]{The Hong Kong University of Science and Technology (Guangzhou).}
\affil[2]{LimX Dynamics Technology Co., Ltd.}
\affil[3]{Shandong University.}
\affil[4]{Institute of Deep Perception Technology, Jiangsu Industrial Technology Research Institute (JITRI).}
\affil[5]{Data61/CSIRO.}
\affil[6]{Griffith University.}

\dates{Draft prepared for arXiv on \today}
\corres{\textsuperscript{\textasteriskcentered} Corresponding author: Yutao Yue.}
\docinfo{\textsuperscript{\ensuremath{\dagger}} Haozhe Jia, Honglei Jin, Yuan Zhang, and Youcheng Fan contributed equally as co-first authors.}

\begin{abstract}
Natural language is an intuitive interface for humanoid robots, yet streaming whole-body control requires control representations that are executable now and anticipatory of future physical transitions. Existing language-conditioned humanoid systems typically generate kinematic references that a low-level tracker must repair reactively, or use latent/action policies whose outputs do not explicitly encode upcoming contact changes, support transfers, and balance preparation. We propose \textbf{DAJI} (\emph{Dynamics-Aligned Joint Intent}), a hierarchical framework that learns an anticipatory joint-intent interface between language generation and closed-loop control. DAJI-Act distills a future-aware teacher into a deployable diffusion action policy through student-driven rollouts, while DAJI-Flow autoregressively generates future intent chunks from language and intent history. Experiments show that DAJI achieves strong results in anticipatory latent learning, single-instruction generation, and streaming instruction following, reaching 94.42\% rollout success on HumanML3D-style generation and 0.152 subsequence FID on BABEL.

\noindent\textbf{\faIcon{globe}\hspace{0.4em}Project Page:} \url{https://hxxxz0.github.io/DAJI_PAGE/}
\end{abstract}


\begin{document}

\normalsize

\maketitle
\thispagestyle{firststyle}

\begin{strip}
    \centering
    \includegraphics[width=0.98\textwidth]{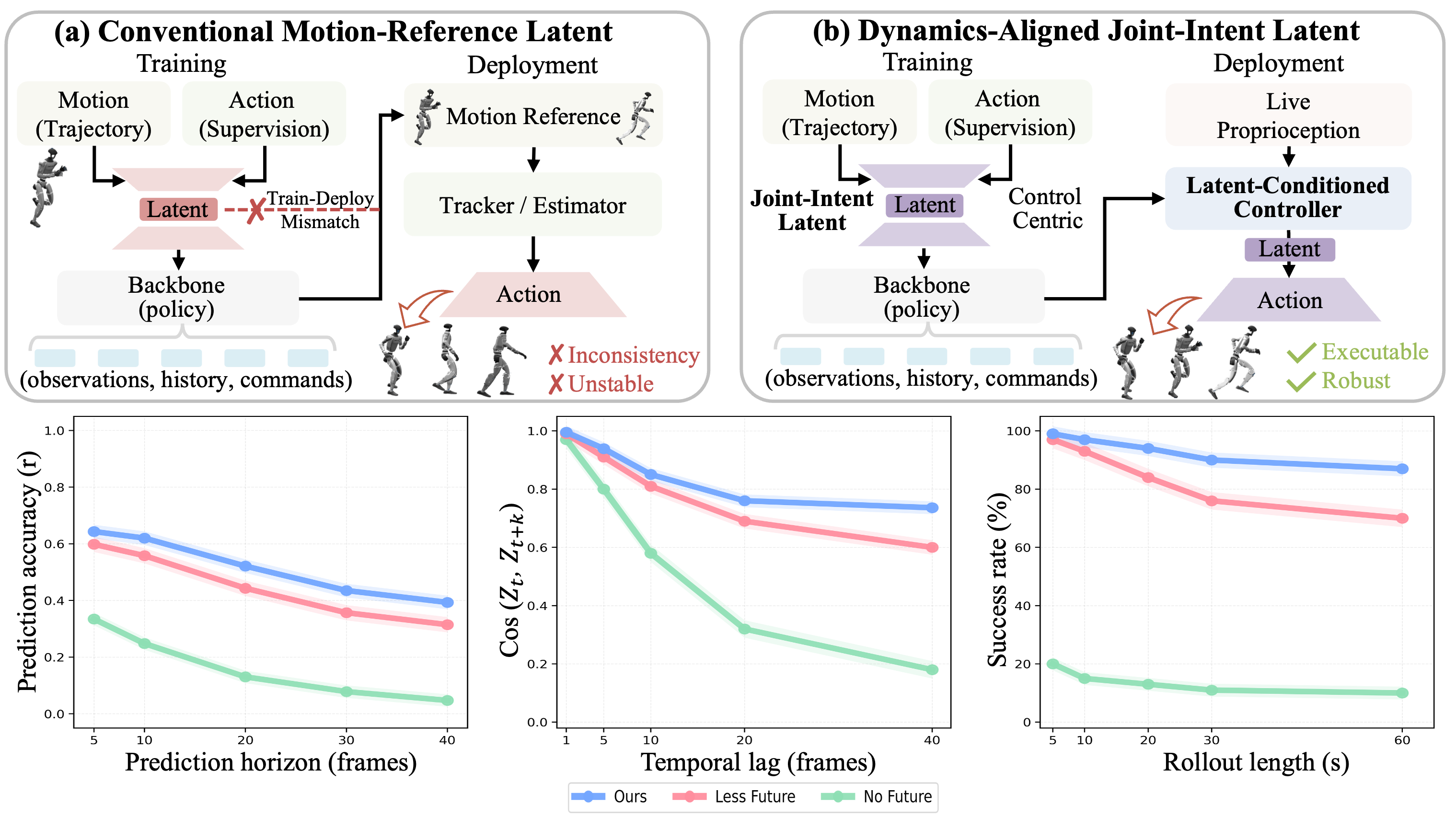}
    \captionof{figure}{
    \textbf{Teaser of DAJI.}
    Instead of using reference trajectories as the deployment interface, DAJI predicts executable and anticipatory joint-intent latents that improve future prediction and long-horizon humanoid control.
    }
    \label{fig:teaser}
\end{strip}

\section{Introduction}

\rhostart{N}atural language is becoming a practical interface for humanoid robots because it allows non-expert users to specify goals, styles, and temporal changes without teleoperation~\cite{ahn2022can, tevet2022mdm}. This paper studies language-conditioned humanoid whole-body control in a streaming setting: given a language instruction and the robot's recent motion history, the system should continuously produce behavior that is semantically aligned, physically executable, and extendable over long horizons. The central question is therefore not only how to generate plausible motion from language, but what intermediate representation should connect language understanding, motion generation, and closed-loop humanoid control~\cite{liang2023code, huang2023voxposer}.

Recent progress has moved this problem beyond offline text-to-motion generation. Diffusion, flow-matching, and autoregressive motion models can synthesize diverse human motions from language~\cite{tevet2022mdm,zhang2023t2mgpt,guo2023momask,barquero2024flowmdm, hu2023motionflow, sato, chen2025freet2mrobusttexttomotiongeneration}, and streaming generators improve online kinematic continuity by conditioning on motion history and incremental text inputs~\cite{xiao2025motionstreamer}. However, these models are mainly optimized in a kinematic space and do not specify how a humanoid robot should realize the motion under contact, balance, actuation, and feedback. Reference-driven humanoid systems address this by generating human or robot-native motion references and executing them with tracking policies~\cite{li2026fromw1,xie2026textop,jia2026echo}. This modular design has enabled impressive demonstrations, but the generated reference remains an external target: when its contact timing, root momentum, support transition, or recovery behavior is inconsistent with the robot's current state, the tracker can only react by compromising either tracking accuracy or stability. Latent-driven or end-to-end approaches further reduce explicit retargeting by coupling language, motion latents, and policies more tightly~\cite{shao2025langwbc,li2025roboghost,wang2025sentinel,yuan2026roboforge,jiang2025uniact}. Yet these latents are often treated as semantic skill codes, compact motion descriptors, or direct action-conditioning variables, rather than as closed-loop motor intents that also express what the body must prepare for next.

This limitation is especially consequential in streaming deployment, where generated chunks are not independent clips: each execution alters the robot's contact state, center of mass, momentum, and feasible continuation set. Consequently, a segment may be locally executable yet fundamentally myopic. Complex maneuvers, such as stepping, turning, or balance recovery require preparatory support shifts and angular-momentum regulation well before the primary motion manifests. Inspired by anticipatory postural adjustment (APA) in human motor control~\cite{massion1992movement,bouisset2008posture}, we argue that the key interface for language-conditioned humanoid control should be an anticipatory motor intent. This interface must remain executable under live proprioception while simultaneously encoding coarse future tendencies across contacts, balance shifts, and task progression.

To this end, we propose \emph{Dynamics-Aligned Joint Intent} (\textbf{DAJI}), a hierarchical framework for streaming language-conditioned humanoid control. At deployment, DAJI replaces motion references with a closed-loop joint-intent interface between semantic generation and control~\cite{peng2018deepmimic, luo2023perpetual}. Instead of generating poses for a tracker, it learns compact intents through controller interaction, unlike decoupled latent skill spaces~\cite{tessler2023calm}. To make these intents anticipatory, a privileged teacher uses multi-horizon future references to capture upcoming support shifts, contact transitions, and balance demands. We then distill \textbf{DAJI-Act} in loop, so the policy is trained under its own closed-loop state distribution. This shapes the latent space around humanoid dynamics and recovery, rather than offline reconstruction alone.

On top of this executable intent space, DAJI trains \textbf{DAJI-Flow}, a language-conditioned flow-matching generator that autoregressively predicts future intent chunks. DAJI-Flow conditions on the language instruction and a compact encoding of recent joint-intent latent history, while scheduled self-conditioning exposes the generator to its own previously generated histories during training. At deployment, each generated joint intent is decoded with live proprioception by \textbf{DAJI-Act}, a lightweight diffusion action policy, into high-frequency joint-position actions. In this way, DAJI decouples semantic generation from feedback control while maintaining a shared representation that is both executable and anticipatory. Experiments show that the learned joint-intent latent contains information about upcoming motion transitions, improves temporal coherence over non-anticipatory alternatives, and leads to better long-horizon closed-loop success. These results support our central claim: for language-conditioned humanoid control, the interface between generation and control should not merely describe the desired motion; it should help prepare the body for what comes next.

Our contributions are summarized as follows:
\begin{itemize}
    \item We propose an anticipatory intent-generation paradigm for streaming language-conditioned humanoid control, addressing the mismatch between language-conditioned semantic planning and closed-loop physical execution through predictive joint-intent generation.

    \item We introduce a dynamics-aligned joint-intent representation that couples immediate closed-loop executability with future motion tendency, enabling the controller to act on both the current command and the upcoming transition.
    
    \item We propose DAJI, a hierarchical generation-to-execution framework that learns joint intents through future-aware privileged control and student-driven in-loop distillation, then generates them with language-conditioned flow matching and scheduled self-conditioning for long-horizon humanoid execution.
\end{itemize}

\section{Related Work}

\paragraph{Reference interfaces for language-driven humanoids.}
Language-conditioned humanoid control commonly converts instructions into motion targets that a robot controller can execute. This line builds on text-to-motion priors and generative models~\cite{guo2022generating,plappert2016kit,tevet2022mdm,zhang2023t2mgpt,guo2023momask,barquero2024flowmdm,hu2023motionflow,Chen_2025,jia2025lumalowdimensionunifiedmotion}. Harmon uses human motion priors and VLM-based refinement to synthesize text-aligned humanoid motions~\cite{jiang2024harmon}. FRoM-W1 generates whole-body human motions and then retargets and tracks them on humanoids~\cite{li2026fromw1}. TextOp makes this pipeline interactive by streaming short-horizon robot-skeleton references~\cite{xie2026textop}, while ECHO improves deployability with a compact robot-native representation between a cloud generator and an edge tracker~\cite{jia2026echo}. These systems show that reference-based modularity is practical. However, the generated command remains a trajectory-like target. When its contact timing, support transition, momentum, or recovery behavior is inconsistent with the robot's current state, the low-level policy must repair the mismatch reactively. DAJI instead uses a controller-distilled joint intent as the generation-control interface, so the command is learned to be executable rather than merely trackable.

\paragraph{Latent and streaming interfaces.}
Recent methods reduce explicit retargeting by coupling language, latents, and whole-body policies more tightly. LangWBC trains an end-to-end language-conditioned policy with a CVAE latent structure~\cite{shao2025langwbc}. RoboGhost conditions a diffusion policy on language-grounded motion latents and bypasses explicit motion decoding~\cite{li2025roboghost}. RoboForge further optimizes latent-driven generation with physical plausibility feedback~\cite{yuan2026roboforge}. SENTINEL directly predicts action chunks from language and proprioception~\cite{wang2025sentinel}, and UniAct uses a shared discrete codebook with a causal streaming pipeline for multimodal humanoid control~\cite{jiang2025uniact}. More broadly, recent representation-centric work has explored structured generative spaces and physics-informed alignment objectives in adjacent domains~\cite{ning2025dctdiffintriguingpropertiesimage,jia2025physicsinformedrepresentationalignmentsparse}. These works move beyond pure reference tracking, but the latent or action interface is not necessarily trained as an anticipatory command under its own closed-loop rollout. MotionStreamer shows that causal latents and self-conditioned autoregression help reduce delay and error accumulation in streaming human motion generation~\cite{xiao2025motionstreamer}. Unlike planner-controller unification strategies that merge planning and control into one model, DAJI keeps a modular hierarchy and learns the interface between the two modules: \textbf{DAJI-Flow} generates future joint intents, while \textbf{DAJI-Act} executes them under closed-loop feedback.

\section{Method}

We present DAJI, a hierarchical framework that maps language instructions to executable latent commands for humanoid whole-body control. The key design is a \emph{joint-intent} interface. Instead of asking a generator to produce kinematic references that a tracker must follow, DAJI generates compact latent intents that are learned through closed-loop control and can be decoded into joint-position actions.

\subsection{Overview}

Given a language instruction $\mathcal{L}$ and recent robot observations, DAJI generates joint-position actions $\mathbf{a}_{1:T}$ over a rollout horizon $T$ while supporting streaming extension. We use $\mathbf{o}_t^{\mathrm{prop}}$ for deployable proprioception, $\mathbf{o}_t^{\mathrm{ref}}$ for training-time reference observations, and $\mathbf{o}_t^{\mathrm{priv}}$ for simulator-only privileged observations unavailable at deployment. Observation details are provided in Appendices~\ref{sec:key_dimensions} and~\ref{sec:obs}.

Instead of generating kinematic references that must be repaired by a low-level tracker, DAJI learns a joint-intent latent $\mathbf{z}_t \in \mathbb{R}^{d_z}$ as the interface between semantic generation and physical execution. This latent is both \emph{executable}, since DAJI-Act decodes it into stable whole-body actions using live proprioception, and \emph{anticipatory}, since it captures coarse information about upcoming contacts, balance shifts, and task progression.

DAJI consists of two modules trained sequentially. \textbf{DAJI-Act} learns the executable joint-intent interface by distilling a future-aware privileged teacher into a deployable diffusion action policy through a stochastic intent bottleneck. During distillation, DAJI-Act drives the simulator itself, so the interface is trained on the student's closed-loop state distribution. \textbf{DAJI-Flow} then learns to autoregressively generate future joint-intent chunks from language and recent intent history. At deployment, DAJI-Flow predicts low-frequency intent chunks, while DAJI-Act decodes each latent with live proprioception at the control frequency.

\subsection{Learning DAJI-Act: Executable Joint-Intent Policy}
\label{sec:joint_intent}

\begin{figure*}[t]
    \centering
    \includegraphics[width=0.98\textwidth]{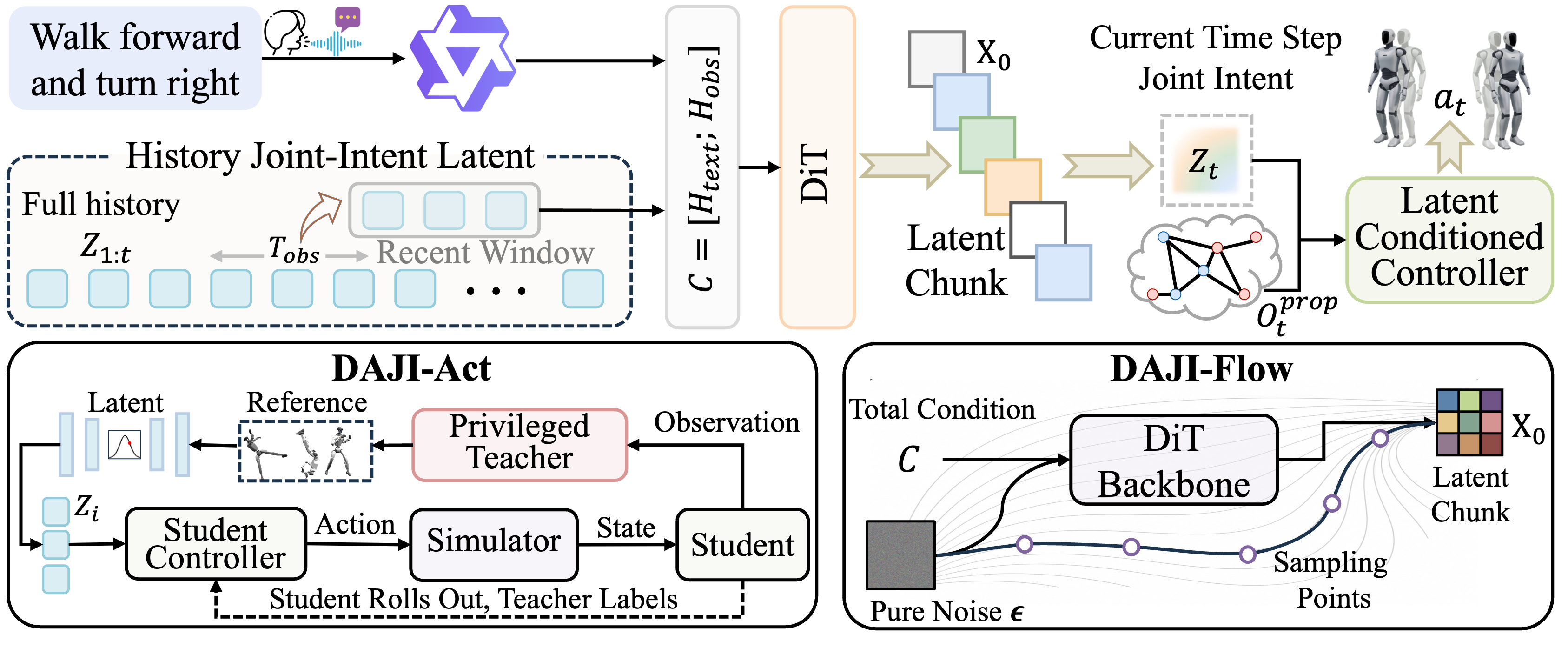}
    \caption{
    \textbf{DAJI framework.}
    DAJI separates online deployment and offline training.
    \textbf{DAJI-Flow} predicts joint-intent latents from language and latent history, while \textbf{DAJI-Act} decodes each intent with live proprioception.
    DAJI-Act learns the executable joint-intent interface through student-driven in-the-loop distillation from a future-aware privileged teacher.
    }
    \label{fig:overview}
    \vspace{-8pt}
\end{figure*}

\begin{figure*}
    \centering
    \includegraphics[width=0.98\textwidth]{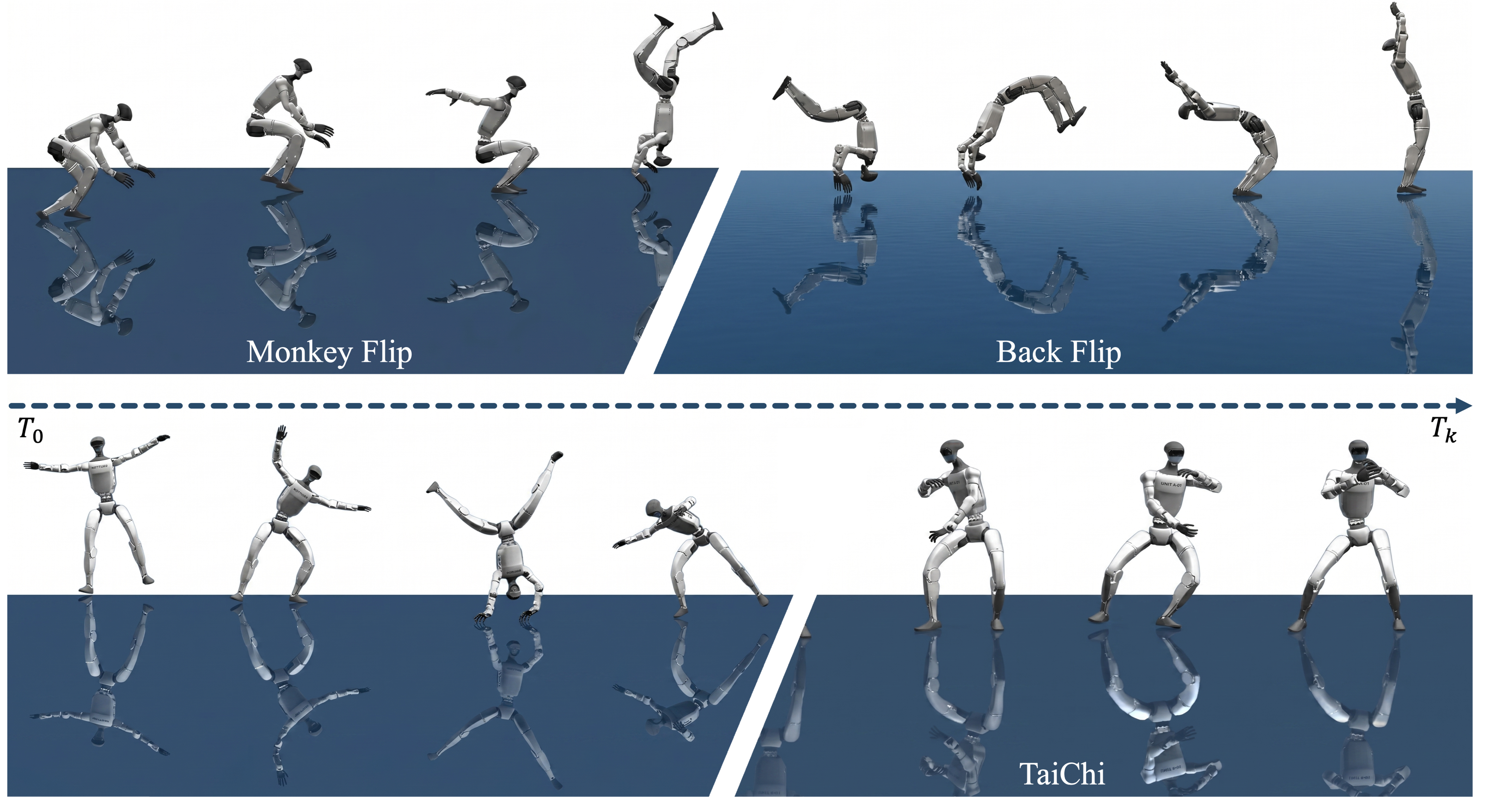}
    \caption{\textbf{Tracker-level rollout visualization in simulation.}
    DAJI decodes generated joint-intent latents into continuous whole-body humanoid motions, including dynamic and highly articulated behaviors.}
    \label{fig:qualitative_tracker_rollout}
\end{figure*}

DAJI-Act is the low-level module that decodes joint-intent latents into robot actions. It uses a stochastic intent bottleneck to compress future-aware reference information and a diffusion action head to imitate teacher actions with live proprioception. Although the bottleneck uses a Gaussian posterior and KL regularization, the latent is optimized as an executable control command rather than a reconstruction target.

\paragraph{Future-aware teacher.}
A locally conditioned controller may track the current pose but fail to prepare for future transitions such as stepping, turning, or balance recovery. We therefore train a privileged teacher that observes not only the current robot state but also multi-horizon reference information. At time $t$, the teacher policy is
\begin{equation}
    \pi_{\mathrm{tea}}
    \bigl(
    \mathbf{a}_t
    \mid
    \mathbf{o}_t^{\mathrm{prop}},
    \mathbf{o}_t^{\mathrm{ref}},
    \mathbf{o}_t^{\mathrm{priv}}
    \bigr),
\end{equation}
where $\mathbf{o}_t^{\mathrm{ref}}$ includes future target frames. The teacher is optimized with PPO~\cite{schulman2017ppo} under whole-body tracking and robustness rewards. Our reward design follows common humanoid motion-tracking objectives and adapts selected tracking and regularization terms from recent compliance-aware whole-body tracking work~\cite{lu2025gentlehumanoid}. Reward definitions, observation groups, symmetry augmentation, and domain randomization are provided in Appendix~\ref{sec:obs}. The teacher is used as a source of future-aware control behavior that will be compressed into the joint-intent latent.

\paragraph{Student-driven in-loop distillation.}
Teacher-driven imitation collects states visited by the teacher, but the deployable policy must act under its own rollout distribution. DAJI therefore performs in-loop distillation on DAJI-Act-induced states. At each step, DAJI-Act decodes a joint-intent-conditioned action and applies it to the simulator. The frozen teacher then evaluates the same state and provides the target action $\mathbf{a}_t^{\mathrm{tea}}$. This trains the intent bottleneck and diffusion action head under the distribution they will encounter at deployment.

DAJI-Act has three parts: a proprioceptive encoder $E_{\mathrm{prop}}$, a stochastic intent encoder $E_{\mathrm{ref}}$, and a diffusion action head. The intent encoder $E_{\mathrm{ref}}$, parameterized by $\psi$, maps $\mathbf{o}_t^{\mathrm{ref}}$ to the stochastic intent distribution
$q_\psi(\mathbf{z}_t|\mathbf{o}_t^{\mathrm{ref}})
= \mathcal{N}(\boldsymbol{\mu}_t,\mathrm{diag}(\boldsymbol{\sigma}_t^2))$,
where $\boldsymbol{\mu}_t$ and $\boldsymbol{\sigma}_t^2$ denote the mean and variance of the intent distribution. We sample the latent by
\begin{equation}
    \mathbf{z}_t
    =
    \boldsymbol{\mu}_t
    +
    \boldsymbol{\sigma}_t \odot \boldsymbol{\epsilon}_z,
    \qquad
    \boldsymbol{\epsilon}_z\sim\mathcal{N}(\mathbf{0},\mathbf{I}).
\end{equation}
The control context is
$\mathbf{c}_t=[E_{\mathrm{prop}}(\mathbf{o}_t^{\mathrm{prop}});\mathbf{z}_t]$.
Given a noised teacher action $\mathbf{x}_\tau$ at diffusion timestep $\tau$,
\begin{equation}
    \mathbf{x}_\tau
    =
    \sqrt{\bar{\alpha}_\tau}\mathbf{a}_t^{\mathrm{tea}}
    +
    \sqrt{1-\bar{\alpha}_\tau}\boldsymbol{\epsilon}_a,
    \qquad
    \boldsymbol{\epsilon}_a\sim\mathcal{N}(\mathbf{0},\mathbf{I}).
\end{equation}
where $\bar{\alpha}_\tau$ is the cumulative noise schedule. Conditioned on $\mathbf{c}_t$, a lightweight denoiser $\mathcal{D}_\phi$ predicts the clean action. The DAJI-Act objective is
\begin{equation}
    \begin{aligned}
    \mathcal{L}_{\mathrm{Act}}
    ={}&
    \mathbb{E}_{(\mathbf{o}_t,\mathbf{a}_t^{\mathrm{tea}})\sim\mathcal{D}_{\mathrm{student}},\,\tau,\,\boldsymbol{\epsilon}_a,\,\boldsymbol{\epsilon}_z}
    \left[
    \left\|
    \mathbf{a}_t^{\mathrm{tea}}
    -
    \mathcal{D}_\phi(\mathbf{x}_\tau,\tau,\mathbf{c}_t)
    \right\|_2^2
    \right] \\
    &\quad+
    \beta\,\mathrm{KL}
    \left(
    q_\psi(\mathbf{z}_t|\mathbf{o}_t^{\mathrm{ref}})
    \,\|\,\mathcal{N}(\mathbf{0},\mathbf{I})
    \right).
    \end{aligned}
\end{equation}
Here, $\beta$ is the KL regularization weight. The first term trains the diffusion action head to imitate future-aware teacher actions on student-visited states. The second term regularizes the stochastic joint-intent bottleneck so that the resulting intent space remains compact and smooth enough for DAJI-Flow. Since the latent is decoded into actions rather than reconstructed references, DAJI-Act should be understood as an executable intent policy rather than a reconstruction module. After distillation, we encode all retargeted reference motions into latent trajectories using the mean intent code from the intent encoder. These text--intent trajectory pairs train DAJI-Flow. The training-only intent encoder is discarded at deployment; DAJI-Act executes generated intents using only live proprioception.

\subsection{Learning DAJI-Flow: Streaming Joint-Intent Generation}
\label{sec:generator}

Once DAJI-Act has learned an executable intent space, language-conditioned humanoid generation becomes the problem of streaming joint-intent latents. \textbf{DAJI-Flow} is a language-conditioned flow-matching DiT that predicts future intent chunks conditioned on the language instruction and recent executed intent history. A frozen language encoder maps $\mathcal{L}$ to text features $\mathbf{H}^{\mathrm{text}}$, and a temporal encoder maps the recent executed latent history to $\mathbf{H}^{\mathrm{obs}}$. The conditioning sequence is
\begin{equation}
    \mathbf{C}=[\mathbf{H}^{\mathrm{text}};\mathbf{H}^{\mathrm{obs}}].
\end{equation}

DAJI-Flow generates the next clean latent chunk $\mathbf{X}_0\in\mathbb{R}^{H\times d_z}$, where $H$ is the chunk length.
Given noise $\boldsymbol{\epsilon}$ and flow time $s\in[0,1]$, we use the path
$\mathbf{X}_s=(1-s)\boldsymbol{\epsilon}+s\mathbf{X}_0$
and train the velocity field with
\begin{equation}
    \mathcal{L}_{\mathrm{Flow}}
    =
    \mathbb{E}_{s,\boldsymbol{\epsilon}}
    \left[
    \left\|
    \mathbf{v}_\theta(\mathbf{X}_s,s,\mathbf{C})
    -
    (\mathbf{X}_0-\boldsymbol{\epsilon})
    \right\|_2^2
    \right].
\end{equation}
At inference, the ODE is integrated for a small number of Euler steps, the resulting chunk is appended to the latent history, and generation continues autoregressively. DAJI-Flow therefore predicts executable future intent tendencies, while DAJI-Act handles high-frequency feedback, perturbation compensation, and contact-level stabilization.

\paragraph{Scheduled self-conditioning.}
Teacher-forced training conditions DAJI-Flow on ground-truth latent histories, whereas deployment conditions it on its own previous predictions. To reduce this exposure bias, we train the generator with multi-chunk self-conditioning. The first chunk is predicted from ground-truth history; subsequent chunks are conditioned on a history buffer augmented with previously generated latent chunks. These generated chunks are detached before being appended to the history buffer. In our default setting, self-conditioning is disabled for the first 100k steps and linearly ramped to probability 1 over the next 40k steps. Details are provided in Appendix~\ref{sec:self_conditioning}.

\subsection{Training and Deployment}

DAJI is trained in three stages. Stage~1 trains the future-aware privileged teacher with PPO. Stage~2 trains \textbf{DAJI-Act} through student-driven in-loop distillation, learning both the stochastic joint-intent bottleneck and the diffusion action head. After convergence, the training-only intent encoder is applied offline to reference motions to construct text--intent trajectory pairs. Stage~3 trains \textbf{DAJI-Flow} with flow matching and scheduled self-conditioning, with the language encoder frozen.

At deployment, each newly received active instruction is encoded once and reused until the next instruction switch. For each chunk, \textbf{DAJI-Flow} predicts future joint-intent latents from the active text embedding and recent intent history. Each chunk contains 15 latent frames, corresponding to 0.3\,s at the 50\,Hz control frequency. Each $\mathbf{z}_t$ is decoded in real time by \textbf{DAJI-Act} conditioned on live proprioception. The privileged teacher and the training-only intent encoder are not used at deployment. DAJI-Flow and DAJI-Act run asynchronously: DAJI-Flow produces low-frequency intent chunks, while DAJI-Act provides low-latency physical execution of each generated latent.

\section{Experiments}
\label{sec:experiments}

\begin{table*}[t]
\centering
\caption{\textbf{Execution interface validation.}
The deployable \textbf{DAJI-Act} policy decodes joint intents with live proprioception and runs in real time on CPU.}
\label{tab:execution_interface}
\CompactTableStyle
\setlength{\tabcolsep}{3.5pt}
\renewcommand{\arraystretch}{1.10}
\begin{tabular*}{\textwidth}{@{\extracolsep{\fill}}lccccc@{}}
\toprule
\textbf{Variant} & \textbf{Succ.} (\%) $\uparrow$ & \textbf{Fall} (\%) $\downarrow$ & \textbf{L-MPJPE} $\downarrow$ & \textbf{Skate} $\downarrow$ & \textbf{Lat. (ms)} $\downarrow$ \\
\midrule
Privileged teacher          & \textbf{92.5}                   & \textbf{7.5}                    & \textbf{29.9}                   & \textbf{131.3}                    & \textbf{1.73} (GPU) \\
DAJI-Act (MLP, 64D)        & \cI \textbf{80.8}               & \cI \textbf{19.2}               & \cI \textbf{39.7}               & \cI \textbf{175.8}                & \cII 4.71 (CPU) \\
DAJI-Act (MLP, 32D)        & \cIII 76.3                      & \cIII 23.7                      & \cII 40.7                       & \cIII 189.5                       & \cIII 4.73 (CPU) \\
DAJI-Act (MLP, 16D)        & \cIV 49.5                       & \cIV 50.5                       & \cIV 58.3                       & \cIV 223.3                        & \cI \textbf{4.28} (CPU) \\
DAJI-Act (Transformer, 64D) & \cII 76.9                       & \cII 23.1                       & \cIII 42.0                      & \cII 178.9                        & \cIV 7.48 (CPU) \\
\bottomrule
\end{tabular*}
\end{table*}

\begin{table*}[t]
\centering
\caption{\textbf{Main comparison on HumanML3D.} $\rightarrow$ means closer to GT is better. The best non-GT result in each metric is bolded, and shading indicates relative rank among non-GT methods.}
\label{tab:humanml3d_main}
\CompactTableStyle
\setlength{\tabcolsep}{3.5pt}
\renewcommand{\arraystretch}{1.10}
\begin{tabular*}{\textwidth}{@{\extracolsep{\fill}}lcccccccc@{}}
\toprule
\textbf{Method} & \textbf{MM-D} $\downarrow$ & \textbf{R@1} $\uparrow$ & \textbf{R@2} $\uparrow$ & \textbf{R@3} $\uparrow$ & \textbf{FID} $\downarrow$ & \textbf{Div} $\rightarrow$ & \textbf{MM} $\uparrow$ & \textbf{Succ. (\%)} $\uparrow$ \\
\midrule
GT & 0.968 & 0.720 & 0.878 & 0.919 & 0.000 & 1.372 & -- & 100.00 \\
GT (Sim) & 1.002 & 0.687 & 0.847 & 0.900 & 0.059 & 1.320 & -- & 94.50 \\
\midrule
LangWBC & 1.462 & 0.205 & 0.328 & 0.412 & 0.510 & 1.045 & 0.365 & 70.50 \\
ECHO & \cV 1.318 & \cV 0.288 & \cV 0.432 & \cV 0.528 & \cV 0.378 & \cV 1.148 & \cV 0.462 & \cV 82.00 \\
FRoM-W1 & 1.392 & 0.245 & 0.380 & 0.471 & 0.445 & 1.095 & 0.412 & 76.00 \\
RoboGhost & \cIV 1.275 & \cIV 0.325 & \cIV 0.468 & \cIV 0.562 & \cIV 0.335 & \cIV 1.178 & \cIV 0.501 & \cIV 85.00 \\
MotionStreamer & \cIII 1.225 & \cIII 0.378 & \cIII 0.542 & \cIII 0.645 & \cIII 0.278 & \cIII 1.205 & \cIII 0.585 & \cIII 87.50 \\
TextOp & \cII 1.175 & \cII 0.425 & \cII 0.595 & \cII 0.698 & \cII 0.225 & \cII 1.228 & \cII 0.628 & \cII 90.00 \\
\midrule
\textbf{DAJI} & \cI \textbf{1.093} & \cI \textbf{0.549} & \cI \textbf{0.711} & \cI \textbf{0.796} & \cI \textbf{0.147} & \cI \textbf{1.291} & \cI \textbf{0.847} & \cI \textbf{94.42} \\
\bottomrule
\end{tabular*}
\end{table*}

\begin{table*}[!t]
\centering
\caption{\textbf{Long-term generation evaluation on BABEL.}
Subsequence metrics evaluate segment-level text-motion quality, while transition metrics evaluate continuity around text-switching boundaries.}
\label{tab:babel_main}
\CompactTableStyle
\setlength{\tabcolsep}{3.5pt}
\renewcommand{\arraystretch}{1.10}
\begin{tabular*}{\textwidth}{@{\extracolsep{\fill}}lcccccccc@{}}
\toprule
\textbf{Method} & \multicolumn{4}{c}{\textbf{Subsequence}} & \multicolumn{4}{c}{\textbf{Transition}} \\
\cmidrule(lr){2-5}\cmidrule(lr){6-9}
& R@3 $\uparrow$ & FID $\downarrow$ & Div $\rightarrow$ & MM-D $\downarrow$
& FID $\downarrow$ & Div $\rightarrow$ & PJ $\downarrow$ & MJ $\downarrow$ \\
\midrule
GT        & 0.654 & 0.000 & 1.299 & 1.299 & 0.000 & 1.182 & 0.043 & 0.015 \\
GT (Sim)  & 0.625 & 0.059 & 1.287 & 1.157 & 0.054 & 1.153 & 0.090 & 0.022 \\
\midrule
TextOp    & 0.287 & 0.538 & 0.985 & 1.514 & 0.594 & 0.784 & 0.231 & 0.156 \\
\midrule
\textbf{DAJI} & \cI \textbf{0.443} & \cI \textbf{0.152} & \cI \textbf{1.219} & \cI \textbf{1.237} & \cI \textbf{0.236} & \cI \textbf{1.033} & \cI \textbf{0.115} & \cI \textbf{0.054} \\
\bottomrule
\end{tabular*}
\end{table*}

We evaluate DAJI as a language-conditioned streaming humanoid control framework. The central question behind our experiments is whether the learned joint-intent latent behaves as an anticipatory command rather than a state-like motion code: it should be executable by the controller, predictive of upcoming motion, and stable when generated autoregressively. We therefore evaluate DAJI along three complementary axes: whether the latent is a viable execution interface, whether its anticipatory temporal structure matters, and whether the resulting generation pipeline improves both single-instruction and streaming language-conditioned humanoid control. Since every generated command must be consumed by a closed-loop humanoid controller, we evaluate DAJI beyond offline language-to-motion metrics and treat rollout executability as a first-class outcome.

\subsection{Experimental Setup}
\label{sec:experimental_setup}

We evaluate DAJI on two complementary settings: a HumanML3D-style robot motion benchmark for single-instruction language-conditioned generation, and BABEL~\cite{punnakkal2021babel} for long-horizon streaming generation with instruction switches. We compare with recent language-conditioned humanoid motion generation and control methods whenever their outputs can be evaluated under the same robot-motion pipeline. Reference-based baselines and GT (Sim) are executed using the same privileged teacher tracker, while DAJI-Act is used only to decode DAJI's joint-intent latents. We report standard text-motion metrics \cite{chen2026towards}, execution success, and transition-continuity metrics. Detailed dataset construction, baseline execution, metric definitions, and default hyperparameters are provided in Appendix~\ref{sec:eval_protocol}.

\subsection{Qualitative Results}
\label{sec:qualitative_results}

\begin{figure*}[!t]
    \centering
    \includegraphics[width=0.98\textwidth]{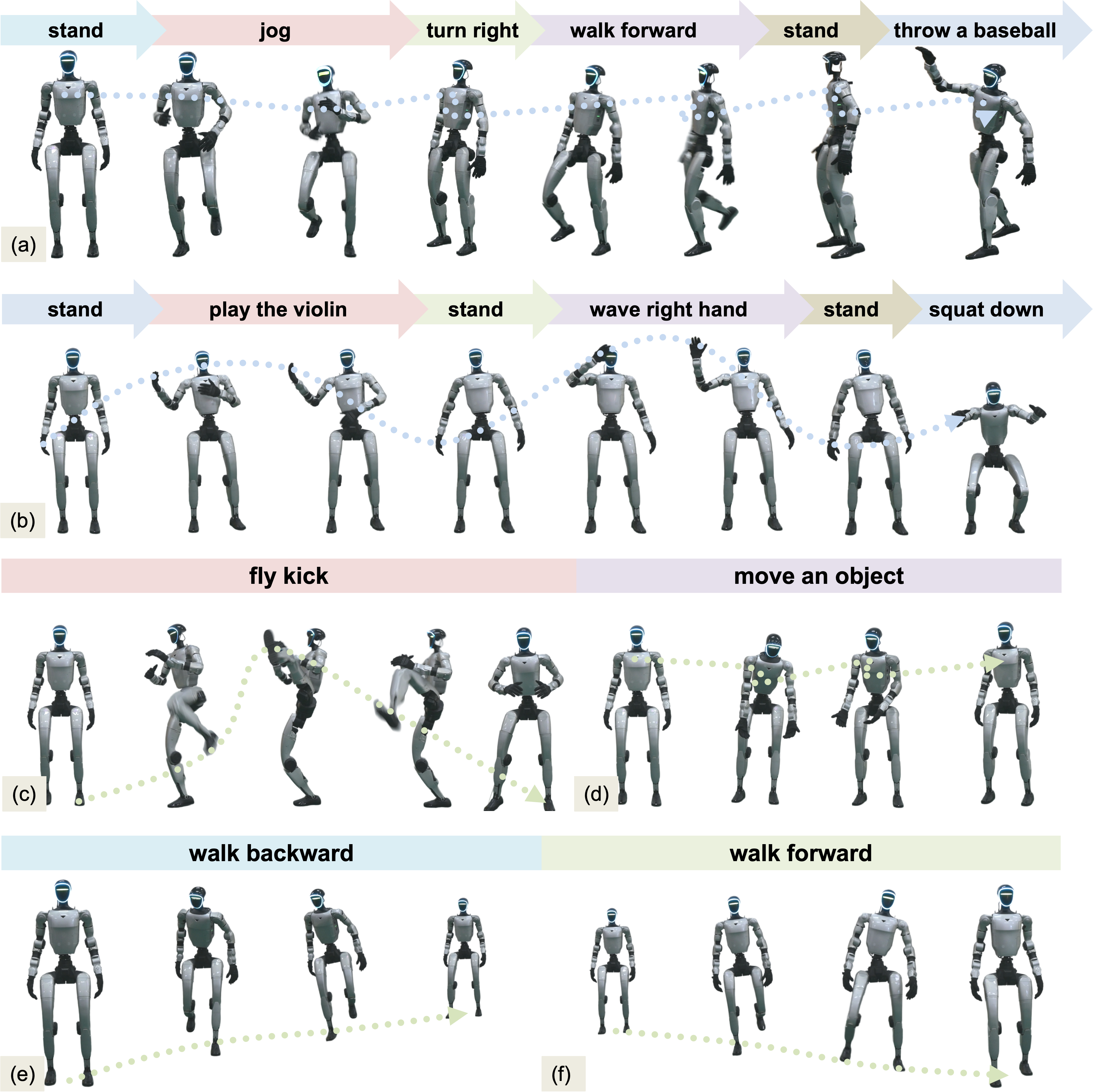}
    \caption{
    \textbf{Qualitative deployment results on physical humanoid hardware.}
    DAJI produces executable motions under both streaming instruction switches and single-instruction generation. Any object-related phrases shown in qualitative prompts are interpreted only as body-motion descriptions; no object state or manipulation outcome is modeled or evaluated.
    }
    \label{fig:qualitative_streaming_deploy}
\end{figure*}

Before quantitative comparisons, we visualize representative deployment results. Figure~\ref{fig:qualitative_streaming_deploy} shows physical hardware deployment, and Figure~\ref{fig:qualitative_tracker_rollout} presents simulation rollouts where the low-level controller decodes generated joint-intent latents into continuous whole-body humanoid motions.

\subsection{Execution Interface Validation}
\label{sec:execution_interface_validation}

We first verify that the learned joint-intent latent is a deployable execution interface rather than an offline representation. The goal here is not to benchmark a universal tracking policy, but to test whether commands expressed in the joint-intent space can be decoded by the deployable controller into stable humanoid behavior under closed-loop dynamics.

Table~\ref{tab:execution_interface} establishes that DAJI's joint-intent latent is not merely useful for generation; it is a viable control interface. The 64D MLP student offers the best practical tradeoff, preserving strong local articulation accuracy and closed-loop stability while remaining real-time on CPU. Reducing the bottleneck to 16D sharply degrades success rate and physical quality, showing that this is not a trivial result of any compact latent. Replacing the 64D MLP with a Transformer also increases latency without improving execution quality, which supports the default controller design used throughout the paper.

\subsection{Main Benchmarks}
\label{sec:main_benchmarks}

\subsubsection{HumanML3D-Style Robot Motion Generation}
\label{sec:humanml3d_eval}

We first evaluate DAJI on the HumanML3D-style robot motion test set, where each sample is paired with a single language instruction. This benchmark measures whether the generator can produce semantically aligned robot motions while preserving executability after export to the humanoid control stack.

Table~\ref{tab:humanml3d_main} shows that DAJI improves both language-motion alignment and rollout executability. For a conservative comparison, reference-based baselines are executed with the same privileged teacher tracker used for GT (Sim), while DAJI-Act only decodes DAJI's joint-intent latents. Thus, the gains mainly reflect the generation--control interface rather than a stronger tracking backend. By generating controller-distilled intents instead of external kinematic references, DAJI produces commands that remain semantically aligned and decodable under feedback control. The gap between GT and GT (Sim) further shows why rollout success should be treated as a first-class metric.

\subsubsection{Streaming Instruction Following on BABEL}
\label{sec:streaming_eval}

We next evaluate DAJI on BABEL, which exposes a different failure mode from single-instruction generation. Here the model must continue from its own latent history while adapting to new instructions at segment boundaries. This directly tests whether the interface remains coherent under long-horizon streaming rollout rather than merely producing one isolated clip.

Table~\ref{tab:babel_main} shows that the advantage of DAJI becomes even clearer in streaming rollout. Compared with the closest compatible streaming baseline, DAJI improves both subsequence quality and boundary continuity. This is exactly the regime where anticipatory intent matters: instruction switches require the policy to prepare support transfer, root momentum, and posture adaptation before the visible transition is complete. Generating in the joint-intent space therefore improves not only semantic alignment but also the smoothness and stability of instruction-conditioned transitions.

\begin{table*}[!t]
\centering
\caption{\textbf{Representation-level latent temporal design ablation.}
Future-aware latents improve predictiveness, coherence, and long-horizon execution.}
\label{tab:latent_design_probe}
\CompactTableStyle
\begin{tabular*}{\textwidth}{@{\extracolsep{\fill}}lcccc@{}}
\toprule
\textbf{Variant} & \textbf{Probe@40} $\uparrow$ & \textbf{Corr.@40} $\uparrow$ & \textbf{Succ.@20s (\%)} $\uparrow$ & \textbf{Succ.@60s (\%)} $\uparrow$ \\
\midrule
No-Future & \cIII 0.047 & \cIII 0.180 & \cIII 13 & \cIII 10 \\
Short-Future & \cII 0.315 & \cII 0.600 & \cII 84 & \cII 70 \\
Full DAJI & \cI \textbf{0.393} & \cI \textbf{0.736} & \cI \textbf{94} & \cI \textbf{87} \\
\bottomrule
\end{tabular*}
\end{table*}

\begin{table*}[!t]
\centering
\caption{\textbf{Generation-level ablation of joint-intent latent temporal design on HumanML3D.}
All variants use the same DAJI-Flow configuration and differ only in the temporal offsets used to construct the DAJI-Act-distilled 64D joint-intent latent. $\rightarrow$ means closer to GT is better.}
\label{tab:latent_design_generation}
\CompactTableStyle
\setlength{\tabcolsep}{3.5pt}
\renewcommand{\arraystretch}{1.10}
\begin{tabular*}{\textwidth}{@{\extracolsep{\fill}}lcccccccc@{}}
\toprule
\textbf{Method} & \textbf{MM-D} $\downarrow$ & \textbf{R@1} $\uparrow$ & \textbf{R@2} $\uparrow$ & \textbf{R@3} $\uparrow$ & \textbf{FID} $\downarrow$ & \textbf{Div} $\rightarrow$ & \textbf{MM} $\uparrow$ & \textbf{Succ. (\%)} $\uparrow$ \\
\midrule
No-Future Latent    & \cIII 1.113 & \cIII 0.502 & \cIII 0.673 & \cIII 0.767 & \cV 0.209   & \cIII 1.279 & \cIV 0.823 & \cV 93.56 \\
Short-Future Latent  & \cIV 1.139  & \cIV 0.466  & \cIV 0.626  & \cIV 0.722  & \cIII 0.180 & \cV 1.250   & \cI \textbf{0.974} & \cIII 93.65 \\
Full DAJI            & \cI \textbf{1.093} & \cI \textbf{0.549} & \cI \textbf{0.711} & \cI \textbf{0.796} & \cI \textbf{0.147} & \cII 1.291 & \cIII 0.847 & \cI \textbf{94.42} \\
Dense-Future Latent  & \cII 1.110  & \cII 0.506  & \cII 0.677  & \cII 0.773  & \cII 0.159  & \cI \textbf{1.294} & \cV 0.697 & \cII 93.88 \\
Long-Future Latent   & \cV 1.148   & \cV 0.436   & \cV 0.595   & \cV 0.697   & \cIV 0.194  & \cIV 1.262 & \cII 0.848 & \cIV 93.61 \\
\bottomrule
\end{tabular*}
\end{table*}

\subsection{Ablations and Design Analysis}
\label{sec:ablations}

We next analyze which design choices make the joint-intent latent an anticipatory command rather than a state-like motion code, and which components stabilize autoregressive generation.

\paragraph{Generator training design.}
We first evaluate the role of scheduled self-conditioning. The DAJI-Flow w/o Self-Conditioning variant removes predicted-history feedback during training and conditions the generator only on ground-truth latent history, corresponding to standard teacher-forced training. Full DAJI instead feeds generated latent chunks back into the history during training, gradually exposing the model to its own induced rollout distribution.

\begin{table}[!t]
\centering
\caption{\textbf{Ablation study of generator-side design on HumanML3D.}}
\label{tab:daji_ablation}
\CompactTableStyle
\setlength{\tabcolsep}{6pt}
\renewcommand{\arraystretch}{1.10}
\begin{tabular*}{0.92\columnwidth}{@{\extracolsep{\fill}}lccc@{}}
\toprule
\textbf{Method} & \textbf{R@3} $\uparrow$ & \textbf{FID} $\downarrow$ & \textbf{Succ.} (\%) $\uparrow$ \\
\midrule
Full DAJI & \cI \textbf{0.796} & \cI \textbf{0.147} & \cI \textbf{94.42} \\
w/o Self-Cond. & 0.667 & 0.153 & 93.29 \\
\bottomrule
\end{tabular*}
\end{table}

Table~\ref{tab:daji_ablation} shows that scheduled self-conditioning improves language alignment under autoregressive rollout, as reflected by the higher R@3. The gain in success rate is modest but consistent, while FID remains comparable. This supports the role of self-conditioning in reducing the train-test mismatch once the model must condition on its own previously generated latent histories.

\paragraph{Joint-intent latent temporal design.}
We next study whether the temporal construction of the joint-intent latent determines whether it behaves as an anticipatory command rather than a reactive state code. This ablation changes only the latent target: for each design, we retrain the DAJI-Act-distilled intent encoder, extract text--latent trajectories, and train the same DAJI-Flow configuration. All variants share the same latent dimensionality, generator, chunk size, self-conditioning schedule, latent-history length, optimizer, data, and metrics. The compared designs differ only in which past and future reference frames are compressed into the latent.

\begin{table}[!t]
\centering
\caption{\textbf{Temporal offset sets for joint-intent latent construction.}}
\label{tab:latent_temporal_offsets}
\CompactTableStyle
\setlength{\tabcolsep}{3pt}
\renewcommand{\arraystretch}{1.12}
\begin{tabularx}{\columnwidth}{@{}l>{\raggedright\arraybackslash}X@{}}
\toprule
\textbf{Variant} & \textbf{Temporal offsets} \\
\midrule
No-Future &
$\{0,-1,-2,-4,-8,-12,-16\}$ \\
Short-Future &
$\{0,+1,+2,-1,-2,-4,-8,-12,-16\}$ \\
Full DAJI &
$\{0,+1,+2,+3,+4,-1,-2,-4,-8,-12,-16\}$ \\
Dense-Future &
$\{0,+1,+2,+3,+4,+5,+6,+7,+8,-1,-2,-4,-8,-12,-16\}$ \\
Long-Future &
$\{0,+1,+2,+3,+4,+8,+12,+16,+20,-1,-2,-4,-8\}$ \\
\bottomrule
\end{tabularx}
\end{table}

Table~\ref{tab:latent_temporal_offsets} summarizes the temporal offset sets used in this ablation. Offset $0$ denotes the current frame, positive offsets denote future reference frames, and negative offsets denote past reference frames.

Table~\ref{tab:latent_design_probe} shows that future-aware latent construction is critical at the representation level. Removing future offsets makes the latent largely reactive, reducing Probe@40 from 0.393 to 0.047 and 60s success from 87\% to 10\%. Short-Future remains functional but is less coherent and less stable over long horizons. Diagnostic definitions for Probe@40 and Corr.@40 are provided in Appendix~\ref{sec:latent_diagnostics}.

Table~\ref{tab:latent_design_generation} shows that the same conclusion carries over to language-conditioned generation. A state-like latent that only summarizes current and past references provides weaker information about upcoming support transfer and body transitions. Conversely, an overly dense or distant future target may contain more motion information but can become harder to infer from text and recent latent history. Dense-Future matches Full DAJI in FID but does not improve semantic alignment or success rate, while Long-Future degrades most generation metrics. The lower fixed-horizon success of No-Future here does not contradict its short-clip HumanML3D export success: the fixed-horizon diagnostic measures extended 20\,s/60\,s closed-loop continuation, whereas the HumanML3D metric evaluates bounded single-instruction clips after export. Taken together, these two tables support the default short-horizon future design as the best tradeoff between anticipatory structure, generative predictability, and closed-loop executability.

\paragraph{Text encoder ablation.}
We further study the effect of the frozen text encoder used by DAJI-Flow. All variants use the same DAJI-Act intent space, DAJI-Flow architecture, training data, chunk size, self-conditioning schedule, and evaluation protocol; only the text encoder is changed.

\begin{table*}[!t]
\centering
\caption{\textbf{Text encoder ablation on the HumanML3D-style benchmark.}
All variants use the same DAJI-Flow configuration and differ only in the frozen text encoder.}
\label{tab:text_encoder_ablation}
\CompactTableStyle
\setlength{\tabcolsep}{3.5pt}
\renewcommand{\arraystretch}{1.10}
\begin{tabular*}{\textwidth}{@{\extracolsep{\fill}}lccccccc@{}}
\toprule
\textbf{Text Encoder} & \textbf{MM-D} $\downarrow$ & \textbf{R@1} $\uparrow$ & \textbf{R@2} $\uparrow$ & \textbf{R@3} $\uparrow$ & \textbf{FID} $\downarrow$ & \textbf{MM} $\uparrow$ & \textbf{Succ. (\%)} $\uparrow$ \\
\midrule
CLIP          & \cIV 1.136 & \cIV 0.470 & \cIV 0.637 & \cIV 0.732 & \cIV 0.258 & \cIV 0.670 & \cIII 93.79 \\
T5            & \cIII 1.109 & \cIII 0.511 & \cII 0.701 & \cII 0.790 & \cIII 0.185 & \cIII 0.756 & \cIV 93.61 \\
Qwen2.5-VL-3B & \cII 1.105 & \cII 0.513 & \cIII 0.683 & \cIII 0.778 & \cII 0.159 & \cI \textbf{0.890} & \cII 94.11 \\
Qwen3-VL-4B   & \cI \textbf{1.093} & \cI \textbf{0.549} & \cI \textbf{0.711} & \cI \textbf{0.796} & \cI \textbf{0.147} & \cII 0.847 & \cI \textbf{94.42} \\
\bottomrule
\end{tabular*}
\end{table*}

Table~\ref{tab:text_encoder_ablation} shows that the choice of text encoder affects language-conditioned intent generation. Qwen3-VL-4B achieves the best semantic alignment metrics, including MM-D and R-precision, as well as the lowest FID and the highest rollout success. Qwen2.5-VL-3B gives slightly higher MultiModality, but Qwen3-VL-4B provides the strongest overall tradeoff. Since the language instruction is encoded only once when it is received or switched, the frozen text encoder is not invoked at every control step. The resulting text features are cached and reused during subsequent latent-chunk generation, so the use of a larger text encoder does not affect the high-frequency DAJI-Act control loop. Hyperparameter sensitivity to self-conditioning depth and latent-history length is reported in Appendix~\ref{sec:hyperparam_sensitivity_appendix}.

\section{Conclusion}
\label{sec:conclusion}

We presented DAJI, a hierarchical framework for streaming language-conditioned humanoid control. At deployment, DAJI replaces motion references with a dynamics-aligned joint-intent representation at the interface between semantic generation and closed-loop control. This representation is executable by a closed-loop controller and anticipatory of upcoming motion transitions. By learning it through future-aware privileged control and student-driven in-loop distillation, then generating it with language-conditioned flow matching, DAJI improves language-motion alignment, temporal coherence, and closed-loop executability on single-instruction and streaming benchmarks. These results suggest that the generation-control interface should not only specify the desired motion, but also help the robot prepare for what comes next.

\clearpage
\printbibliography

@inproceedings{guo2022generating,
  title={Generating Diverse and Natural 3D Human Motions from Text},
  author={Guo, Chuan and Zou, Shihao and Zuo, Xinxin and Wang, Sen and Ji, Wei and Li, Xingyu and Cheng, Li},
  booktitle={CVPR},
  year={2022}
}

@article{plappert2016kit,
  title={The KIT Motion-Language Dataset},
  author={Plappert, Matthias and Mandery, Christian and Asfour, Tamim},
  journal={Big Data},
  volume={4},
  number={4},
  pages={236--252},
  year={2016}
}

@article{tevet2022mdm,
  title={Human Motion Diffusion Model},
  author={Tevet, Guy and Raab, Sigal and Gordon, Brian and Shafir, Yonatan and Cohen-Or, Daniel and Bermano, Amit H.},
  journal={arXiv preprint arXiv:2209.14916},
  year={2022}
}

@inproceedings{zhang2023t2mgpt,
  title={T2M-GPT: Generating Human Motion from Textual Descriptions with Discrete Representations},
  author={Zhang, Jianrong and Zhang, Yangsong and Cun, Xiaodong and Huang, Shaoli and Zhang, Yong and Zhao, Hongwei and Lu, Hongtao and Shen, Xi},
  booktitle={CVPR},
  year={2023}
}

@article{guo2023momask,
  title={MoMask: Generative Masked Modeling of 3D Human Motions},
  author={Guo, Chuan and Mu, Yuxuan and Javed, Muhammad Gohar and Wang, Sen and Cheng, Li},
  journal={arXiv preprint arXiv:2312.00063},
  year={2023}
}

@inproceedings{barquero2024flowmdm,
  title={Seamless Human Motion Composition with Blended Positional Encodings},
  author={Barquero, German and Escalera, Sergio and Palmero, Cristina},
  booktitle={CVPR},
  year={2024}
}

@article{hu2023motionflow,
  title={Motion Flow Matching for Human Motion Synthesis and Editing},
  author={Hu, Vincent Tao and Yin, Wenzhe and Ma, Pingchuan and Chen, Yunlu and Fernando, Basura and Asano, Yuki M. and Gavves, Efstratios and Mettes, Pascal and Ommer, Bjorn and Snoek, Cees G. M.},
  journal={arXiv preprint arXiv:2312.08895},
  year={2023}
}

@inproceedings{jiang2024harmon,
  title={Harmon: Whole-Body Motion Generation of Humanoid Robots from Language Descriptions},
  author={Jiang, Zhenyu and Xie, Yuqi and Li, Jinhan and Yuan, Ye and Zhu, Yifeng and Zhu, Yuke},
  booktitle={Conference on Robot Learning},
  year={2024},
  eprint={2410.12773},
  archivePrefix={arXiv},
  primaryClass={cs.RO}
}

@misc{li2026fromw1,
  title={{FRoM-W1}: Towards General Humanoid Whole-Body Control with Language Instructions},
  author={Li, Peng and Zhuang, Zihan and Gao, Yangfan and Dong, Yi and Li, Sixian and Jiang, Changhao and Dou, Shihan and Xi, Zhiheng and Zhou, Enyu and Huang, Jixuan and Li, Hui and Gong, Jingjing and Ma, Xingjun and Gui, Tao and Wu, Zuxuan and Zhang, Qi and Huang, Xuanjing and Jiang, Yu-Gang and Qiu, Xipeng},
  year={2026},
  eprint={2601.12799},
  archivePrefix={arXiv},
  primaryClass={cs.RO}
}

@misc{xie2026textop,
  title={{TextOp}: Real-time Interactive Text-Driven Humanoid Robot Motion Generation and Control},
  author={Xie, Weiji and Zheng, Jiakun and Han, Jinrui and Shi, Jiyuan and Zhang, Weinan and Bai, Chenjia and Li, Xuelong},
  year={2026},
  eprint={2602.07439},
  archivePrefix={arXiv},
  primaryClass={cs.RO}
}

@misc{jia2026echo,
  title={{ECHO}: Edge-Cloud Humanoid Orchestration for Language-to-Motion Control},
  author={Jia, Haozhe and Song, Jianfei and Zhang, Yuan and Jin, Honglei and Fan, Youcheng and Chen, Wenshuo and Zhang, Wei and Yue, Yutao},
  year={2026},
  eprint={2603.16188},
  archivePrefix={arXiv},
  primaryClass={cs.CV}
}

@misc{shao2025langwbc,
  title={{LangWBC}: Language-directed Humanoid Whole-Body Control via End-to-end Learning},
  author={Shao, Yiyang and Huang, Xiaoyu and Zhang, Bike and Liao, Qiayuan and Gao, Yuman and Chi, Yufeng and Li, Zhongyu and Shao, Sophia and Sreenath, Koushil},
  year={2025},
  eprint={2504.21738},
  archivePrefix={arXiv},
  primaryClass={cs.RO}
}

@misc{li2025roboghost,
  title={From Language to Locomotion: Retargeting-free Humanoid Control via Motion Latent Guidance},
  author={Li, Zhe and Chi, Cheng and Wei, Yangyang and Zhu, Boan and Peng, Yibo and Huang, Tao and Wang, Pengwei and Wang, Zhongyuan and Zhang, Shanghang and Xu, Chang},
  year={2025},
  eprint={2510.14952},
  archivePrefix={arXiv},
  primaryClass={cs.RO}
}

@misc{yuan2026roboforge,
  title={{RoboForge}: Physically Optimized Text-guided Whole-Body Locomotion for Humanoids},
  author={Yuan, Xichen and Li, Zhe and Lyu, Bofan and Zuo, Kuangji and Lu, Yanshuo and Li, Gen and Yang, Jianfei},
  year={2026},
  eprint={2603.17927},
  archivePrefix={arXiv},
  primaryClass={cs.RO}
}

@misc{wang2025sentinel,
  title={{SENTINEL}: A Fully End-to-End Language-Action Model for Humanoid Whole Body Control},
  author={Wang, Yuxuan and Jiang, Haobin and Yao, Shiqing and Ding, Ziluo and Lu, Zongqing},
  year={2025},
  eprint={2511.19236},
  archivePrefix={arXiv},
  primaryClass={cs.RO}
}

@misc{jiang2025uniact,
  title={{UniAct}: Unified Motion Generation and Action Streaming for Humanoid Robots},
  author={Jiang, Nan and He, Zimo and Yu, Wanhe and Pang, Lexi and Li, Yunhao and Li, Hongjie and Cui, Jieming and Li, Yuhan and Wang, Yizhou and Zhu, Yixin and Huang, Siyuan},
  year={2025},
  eprint={2512.24321},
  archivePrefix={arXiv},
  primaryClass={cs.CV}
}

@inproceedings{xiao2025motionstreamer,
  title={{MotionStreamer}: Streaming Motion Generation via Diffusion-based Autoregressive Model in Causal Latent Space},
  author={Xiao, Lixing and Lu, Shunlin and Pi, Huaijin and Fan, Ke and Pan, Liang and Zhou, Yueer and Feng, Ziyong and Zhou, Xiaowei and Peng, Sida and Wang, Jingbo},
  booktitle={Proceedings of the IEEE/CVF International Conference on Computer Vision},
  year={2025},
  eprint={2503.15451},
  archivePrefix={arXiv},
  primaryClass={cs.CV}
}

@article{massion1992movement,
  title={Movement, posture and equilibrium: interaction and coordination},
  author={Massion, Jean},
  journal={Progress in neurobiology},
  volume={38},
  number={1},
  pages={35--56},
  year={1992},
  publisher={Elsevier}
}

@article{bouisset2008posture,
  title={Posture, dynamic stability, and voluntary movement},
  author={Bouisset, Simon and Do, Manh-Cuong},
  journal={Neurophysiologie Clinique/Clinical Neurophysiology},
  volume={38},
  number={6},
  pages={345--362},
  year={2008},
  publisher={Elsevier}
}

@article{peng2018deepmimic,
  title={Deepmimic: Example-guided deep reinforcement learning of physics-based character skills},
  author={Peng, Xue Bin and Abbeel, Pieter and Levine, Sergey and Van de Panne, Michiel},
  journal={ACM Transactions On Graphics (TOG)},
  volume={37},
  number={4},
  pages={1--14},
  year={2018},
  publisher={ACM New York, NY, USA}
}

@inproceedings{luo2023perpetual,
  title={Perpetual humanoid control for real-time simulated avatars},
  author={Luo, Zhengyi and Cao, Jinkun and Kitani, Kris and Xu, Weipeng and others},
  booktitle={Proceedings of the IEEE/CVF International Conference on Computer Vision},
  pages={10895--10904},
  year={2023}
}

@inproceedings{tessler2023calm,
  title={Calm: Conditional adversarial latent models for directable virtual characters},
  author={Tessler, Chen and Kasten, Yoni and Guo, Yunrong and Mannor, Shie and Chechik, Gal and Peng, Xue Bin},
  booktitle={ACM SIGGRAPH 2023 conference proceedings},
  pages={1--9},
  year={2023}
}

@article{ahn2022can,
  title={Do as i can, not as i say: Grounding language in robotic affordances},
  author={Ahn, Michael and Brohan, Anthony and Brown, Noah and Chebotar, Yevgen and Cortes, Omar and David, Byron and Finn, Chelsea and Fu, Chuyuan and Gopalakrishnan, Keerthana and Hausman, Karol and others},
  journal={arXiv preprint arXiv:2204.01691},
  year={2022}
}

@inproceedings{liang2023code,
  title={Code as policies: Language model programs for embodied control},
  author={Liang, Jacky and Huang, Wenlong and Xia, Fei and Xu, Peng and Hausman, Karol and Ichter, Brian and Florence, Pete and Zeng, Andy},
  booktitle={2023 IEEE International conference on robotics and automation (ICRA)},
  pages={9493--9500},
  year={2023},
  organization={IEEE}
}

@article{huang2023voxposer,
  title={Voxposer: Composable 3d value maps for robotic manipulation with language models},
  author={Huang, Wenlong and Wang, Chen and Zhang, Ruohan and Li, Yunzhu and Wu, Jiajun and Fei-Fei, Li},
  journal={arXiv preprint arXiv:2307.05973},
  year={2023}
}

@InProceedings{punnakkal2021babel,
  author    = {Punnakkal, Abhinanda R. and Chandrasekaran, Arjun and Athanasiou, Nikos and Quiros-Ramirez, Alejandra and Black, Michael J.},
  title     = {BABEL: Bodies, Action and Behavior With English Labels},
  booktitle = {Proceedings of the IEEE/CVF Conference on Computer Vision and Pattern Recognition (CVPR)},
  month     = {June},
  year      = {2021},
  pages     = {722--731}
}

@article{schulman2017ppo,
  author      = {Schulman, John and Wolski, Filip and Dhariwal, Prafulla and Radford, Alec and Klimov, Oleg},
  title       = {Proximal Policy Optimization Algorithms},
  journal     = {CoRR},
  volume      = {abs/1707.06347},
  year        = {2017},
  url         = {http://arxiv.org/abs/1707.06347},
  eprinttype  = {arXiv},
  eprint      = {1707.06347}
}

@inproceedings{ho2020ddpm,
  author    = {Ho, Jonathan and Jain, Ajay and Abbeel, Pieter},
  title     = {Denoising Diffusion Probabilistic Models},
  booktitle = {Advances in Neural Information Processing Systems},
  volume    = {33},
  pages     = {6840--6851},
  publisher = {Curran Associates, Inc.},
  year      = {2020},
  url       = {https://proceedings.neurips.cc/paper/2020/file/4c5bcfec8584af0d967f1ab10179ca4b-Paper.pdf}
}

@inproceedings{song2021ddim,
  author    = {Song, Jiaming and Meng, Chenlin and Ermon, Stefano},
  title     = {Denoising Diffusion Implicit Models},
  booktitle = {9th International Conference on Learning Representations, ICLR 2021},
  publisher = {OpenReview.net},
  year      = {2021},
  url       = {https://openreview.net/forum?id=St1giarCHLP}
}

@inproceedings{lipman2023flow,
  author    = {Lipman, Yaron and Chen, Ricky T. Q. and Ben-Hamu, Heli and Nickel, Maximilian and Le, Matthew},
  title     = {Flow Matching for Generative Modeling},
  booktitle = {The Eleventh International Conference on Learning Representations, ICLR 2023},
  publisher = {OpenReview.net},
  year      = {2023},
  url       = {https://openreview.net/forum?id=PqvMRDCJT9t}
}

@InProceedings{peebles2023dit,
  author    = {Peebles, William and Xie, Saining},
  title     = {Scalable Diffusion Models with Transformers},
  booktitle = {Proceedings of the IEEE/CVF International Conference on Computer Vision (ICCV)},
  month     = {October},
  year      = {2023},
  pages     = {4195--4205}
}

@inproceedings{todorov2012mujoco,
  author    = {Todorov, Emanuel and Erez, Tom and Tassa, Yuval},
  title     = {MuJoCo: A Physics Engine for Model-Based Control},
  booktitle = {2012 IEEE/RSJ International Conference on Intelligent Robots and Systems},
  pages     = {5026--5033},
  year      = {2012},
  doi       = {10.1109/IROS.2012.6386109},
  publisher = {IEEE}
}

@article{Bai2025Qwen3VLTR,
  title={Qwen3-VL Technical Report},
  author={Bai, Shuai and Cai, Yuxuan and Chen, Ruizhe and Chen, Keqin and others},
  journal={arXiv preprint arXiv:2511.21631},
  year={2025},
  url={https://arxiv.org/abs/2511.21631}
}

@misc{ning2025dctdiffintriguingpropertiesimage,
  title         = {DCTdiff: Intriguing Properties of Image Generative Modeling in the DCT Space},
  author        = {Ning, Mang and Li, Mingxiao and Su, Jianlin and Jia, Haozhe and Liu, Lanmiao and Bene{\v s}, Martin and Chen, Wenshuo and Salah, Albert Ali and Ertugrul, Itir Onal},
  year          = {2025},
  eprint        = {2412.15032},
  archivePrefix = {arXiv},
  primaryClass  = {cs.CV},
  url           = {https://arxiv.org/abs/2412.15032}
}

@misc{jia2025physicsinformedrepresentationalignmentsparse,
  title         = {Physics-Informed Representation Alignment for Sparse Radio-Map Reconstruction},
  author        = {Jia, Haozhe and Chen, Wenshuo and Huang, Zhihui and Wang, Lei and Xiao, Hongru and Jia, Nanqian and Wu, Keming and Lai, Songning and Tian, Bowen and Yue, Yutao},
  year          = {2025},
  eprint        = {2501.19160},
  archivePrefix = {arXiv},
  primaryClass  = {cs.CV},
  url           = {https://arxiv.org/abs/2501.19160}
}

@inproceedings{Chen_2025,
  author    = {Chen, Wenshuo and Yu, Kuimou and Jia, Haozhe and Yuan, Kaishen and Huang, Zexu and Tian, Bowen and Lai, Songning and Xiao, Hongru and Zhang, Erhang and Wang, Lei and Yue, Yutao},
  title     = {ANT: Adaptive Neural Temporal-Aware Text-to-Motion Model},
  booktitle = {Proceedings of the 33rd ACM International Conference on Multimedia},
  series    = {MM '25},
  publisher = {ACM},
  year      = {2025},
  month     = oct,
  pages     = {9852--9861},
  doi       = {10.1145/3746027.3755168},
  url       = {http://dx.doi.org/10.1145/3746027.3755168}
}

@misc{jia2025lumalowdimensionunifiedmotion,
  title         = {LUMA: Low-Dimension Unified Motion Alignment with Dual-Path Anchoring for Text-to-Motion Diffusion Model},
  author        = {Jia, Haozhe and Chen, Wenshuo and Lin, Yuqi and Yang, Yang and Wang, Lei and Ning, Mang and Tian, Bowen and Lai, Songning and Jia, Nanqian and Chen, Yifan and Yue, Yutao},
  year          = {2025},
  eprint        = {2509.25304},
  archivePrefix = {arXiv},
  primaryClass  = {cs.CV},
  url           = {https://arxiv.org/abs/2509.25304}
}

@misc{lu2025gentlehumanoid,
      title={GentleHumanoid: Learning Upper-body Compliance for Contact-rich Human and Object Interaction}, 
      author={Qingzhou Lu and Yao Feng and Baiyu Shi and Michael Piseno and Zhenan Bao and C. Karen Liu},
      year={2025},
      eprint={2511.04679},
      archivePrefix={arXiv},
      primaryClass={cs.RO},
      url={https://arxiv.org/abs/2511.04679}, 
}

@inproceedings{chen2026towards,
  title={Towards Better Evaluation Metrics for Text-to-Motion Generation},
  author={Chen, Wenshuo and Jia, Haozhe and Yu, Kuimou and Lai, Songning and Wang, Lei and Yue, Yutao},
  booktitle={The Second International Workshop on Transformative Insights in Multifaceted Evaluation at The Web Conference 2026}
}

@misc{chen2025freet2mrobusttexttomotiongeneration,
      title={Free-T2M: Robust Text-to-Motion Generation for Humanoid Robots via Frequency-Domain}, 
      author={Wenshuo Chen and Haozhe Jia and Songning Lai and Lei Wang and Yuqi Lin and Hongru Xiao and Lijie Hu and Yutao Yue},
      year={2025},
      eprint={2501.18232},
      archivePrefix={arXiv},
      primaryClass={cs.CV},
      url={https://arxiv.org/abs/2501.18232}, 
}

@inproceedings{10.1145/3746027.3755441,
author = {Tian, Bowen and Chen, Wenshuo and Li, Zexi and Lai, Songning and Wu, Jiemin and Yue, Yutao},
title = {Text2Weight: Bridging Natural Language and Neural Network Weight Spaces},
year = {2025},
isbn = {9798400720352},
publisher = {Association for Computing Machinery},
address = {New York, NY, USA},
url = {https://doi.org/10.1145/3746027.3755441},
doi = {10.1145/3746027.3755441},
booktitle = {Proceedings of the 33rd ACM International Conference on Multimedia},
pages = {10152–10160},
numpages = {9},
location = {Dublin, Ireland},
series = {MM '25}
}

@misc{li2026deltascoremattersspatial,
      title={Delta Score Matters! Spatial Adaptive Multi Guidance in Diffusion Models},
      author={Haosen Li and Wenshuo Chen and Lei Wang and Shaofeng Liang and Bowen Tian and Soning Lai and Yutao Yue},
      year={2026},
      eprint={2604.26503},
      archivePrefix={arXiv},
      primaryClass={cs.CV},
      url={https://arxiv.org/abs/2604.26503}, 
}

@misc{li2026oraclenoisefastersemantic,
      title={Oracle Noise: Faster Semantic Spherical Alignment for Interpretable Latent Optimization},
      author={Haosen Li and Wenshuo Chen and Lei Wang and Shaofeng Liang and Haozhe Jia and Yutao Yue},
      year={2026},
      eprint={2604.23540},
      archivePrefix={arXiv},
      primaryClass={cs.CV},
      url={https://arxiv.org/abs/2604.23540}, 
}

@misc{li2026z2samplingzerocostzigzagtrajectories,
      title={$Z^2$-Sampling: Zero-Cost Zigzag Trajectories for Semantic Alignment in Diffusion Models},
      author={Haosen Li and Wenshuo Chen and Shaofeng Liang and Lei Wang and Kaishen Yuan and Yutao Yue},
      year={2026},
      eprint={2604.23536},
      archivePrefix={arXiv},
      primaryClass={cs.CV},
      url={https://arxiv.org/abs/2604.23536}, 
}

@misc{chen2025polarisprojectionorthogonalsquaresrobust,
      title={POLARIS: Projection-Orthogonal Least Squares for Robust and Adaptive Inversion in Diffusion Models},
      author={Wenshuo Chen and Haosen Li and Shaofeng Liang and Lei Wang and Haozhe Jia and Kaishen Yuan and Jieming Wu and Bowen Tian and Yutao Yue},
      year={2025},
      eprint={2512.00369},
      archivePrefix={arXiv},
      primaryClass={cs.CV},
      url={https://arxiv.org/abs/2512.00369}, 
}

@inproceedings{10.1145/3746027.3758161,
author = {Jia, Haozhe and Chen, Wenshuo and Huang, Zhihui and Wang, Lei and Xiao, Hongru and Jia, Nanqian and Wu, Keming and Lai, Songning and Tian, Bowen and Yue, Yutao},
title = {Physics-Informed Representation Alignment for Sparse Radio-Map Reconstruction},
year = {2025},
isbn = {9798400720352},
publisher = {Association for Computing Machinery},
address = {New York, NY, USA},
url = {https://doi.org/10.1145/3746027.3758161},
doi = {10.1145/3746027.3758161},
booktitle = {Proceedings of the 33rd ACM International Conference on Multimedia},
pages = {12352–12360},
numpages = {9},
location = {Dublin, Ireland},
series = {MM '25}
}

@inproceedings{sato,
author = {chen, Wenshuo and Xiao, Hongru and Zhang, Erhang and Hu, Lijie and Wang, Lei and Liu, Mengyuan and Chen, Chen},
title = {SATO: Stable Text-to-Motion Framework},
year = {2024},
isbn = {9798400706868},
publisher = {Association for Computing Machinery},
address = {New York, NY, USA},
url = {https://doi.org/10.1145/3664647.3681034},
doi = {10.1145/3664647.3681034},
booktitle = {Proceedings of the 32nd ACM International Conference on Multimedia},
pages = {6989–6997},
numpages = {9},
location = {Melbourne VIC, Australia},
series = {MM '24}
}

\clearpage
\onecolumn
\appendix
\section{Evaluation Protocol and Experimental Setup Details}
\label{sec:eval_protocol}
\label{sec:experimental_setup_details}

This section first summarizes the datasets, baselines, metrics, and default configuration used throughout the paper, then gives the evaluator construction and the detailed HumanML3D-style and BABEL evaluation protocols.

\subsection{Datasets, Baselines, and Metrics}

We evaluate DAJI on two complementary settings. The HumanML3D-style robot motion benchmark measures single-instruction language-conditioned generation using language annotations paired with robot motion trajectories. BABEL~\cite{punnakkal2021babel} evaluates long-horizon streaming generation with instruction switches. We use frame-level action annotations to construct text streams, generate each labeled segment for its original duration, and append the generated segment to the latent history before generating the next one. This enables evaluation of both segment-level semantic alignment and transition quality around instruction boundaries.

We compare with recent language-conditioned humanoid motion generation and control methods, including LangWBC, ECHO, TextOp, FRoM-W1, RoboGhost and MotionStreamer, whenever their outputs can be evaluated under the same robot-motion pipeline. Methods that output human motions or kinematic references are converted into the common robot-motion representation and executed with the same privileged teacher tracker used for GT (Sim). DAJI-Act is used only to decode DAJI's joint-intent latents, not external kinematic references. For policy-based methods, we evaluate executed robot trajectories when available. TextOp is the closest compatible streaming reference-based baseline on BABEL. This protocol gives reference-based baselines a strong tracking backend and is therefore conservative with respect to DAJI's execution advantage, but the resulting success rates should still be interpreted as end-to-end interface validity rather than as a pure low-level controller comparison.

We report standard text-motion metrics, including MM-D, R-precision at top 1/2/3, FID, Diversity, and MultiModality. For closed-loop execution, we report success rate, fall rate, local-frame MPJPE, foot skating, and latency. On HumanML3D, we additionally report rollout success after exporting generated motions to the robot-motion execution pipeline. On BABEL, we report subsequence-level generation metrics and transition-level continuity metrics, where transition FID and Diversity are computed on boundary-centered clips and smoothness is measured by peak jerk (PJ) and mean jerk (MJ). For anticipatory analysis, we further use future-prediction probes, latent temporal correlation, and long-horizon rollout success.

Unless otherwise specified, DAJI uses eight self-conditioned chunks during training and conditions on $K=8$ recent joint-intent chunks during inference. Each chunk contains 15 frames. The self-conditioning depth controls how many predicted chunks are fed back during training, while $K$ controls the latent-history length observed by the generator.

\paragraph{Compute resources and precision.}
Teacher tracker and DAJI-Act training are conducted on 8 NVIDIA A100 GPUs, while DAJI-Flow generator training is conducted on 4 NVIDIA A100 GPUs. Unless otherwise specified, the frozen VLM runs in bfloat16 and the DiT generator head is trained in fp32. Reported controller latency is measured separately on CPU or GPU, as indicated in the corresponding tables.

\subsection{Evaluator}

We use a text-motion evaluator to compute MM-D, R-precision, FID, Diversity, and MultiModality. The evaluator contains a text encoder and a motion encoder, which map language descriptions and robot motion clips into a shared embedding space. In our implementation, MM-D corresponds to the matching score computed by the evaluator, i.e., the average Euclidean distance between paired text and motion embeddings. R-precision is computed by retrieving the matched motion from a batch of candidate motions using the text embedding as query. FID, Diversity, and MultiModality are computed in the evaluator motion-feature space.

The evaluator is trained on paired text-motion data using a contrastive objective. Matched text-motion pairs are pulled closer in the embedding space, while mismatched pairs in the same batch are pushed apart. For HumanML3D-style evaluation, the evaluator is trained on paired text descriptions and robot motion clips from the HumanML3D-style robot motion dataset. For BABEL evaluation, we construct text-motion pairs from frame-level action annotations, where each annotated segment provides a text label and a corresponding robot motion segment. The evaluator is frozen after training and is used only for evaluation.

\subsection{HumanML3D-style Evaluation}

The HumanML3D-style benchmark contains language annotations paired with robot motion trajectories converted into our robot-motion representation. For each test prompt, the generator autoregressively predicts a sequence of 64-dimensional joint-intent latent chunks, which are then exported to robot motion for evaluation.

We filter out motions shorter than 100 frames before evaluation. Motions are padded or truncated to a maximum length of 490 frames. We report two reference rows in the main table. GT denotes the original dataset motions. GT (Sim) denotes the trajectories obtained after passing the corresponding dataset motions through the same robot-motion export and simulation validation pipeline used for generated motions. The gap between GT and GT (Sim) reflects the degradation introduced by robot-motion export, tracking, and simulation validation, rather than errors from the text-conditioned generator.

For generated methods, rollout success is computed over all generated samples. Text-motion metrics are computed only on successfully exported motions, while success rate is reported separately. This avoids hiding invalid generations inside text-motion metrics, so these metrics should be interpreted jointly with success rate rather than in isolation.

\subsection{Robot-Motion Execution and Validation}

GT (Sim) is obtained by executing the dataset motion references in simulation with the privileged teacher tracker. For reference-based baselines that follow a generator--reference--tracker pipeline, we first convert their generated references into the common robot-motion representation and then execute them with the same teacher tracker. DAJI-Act is used only to decode DAJI's joint-intent latents and is not used to execute external kinematic references. This keeps the execution backend consistent for GT (Sim) and reference-based baselines, so rollout success reflects the quality of the generated references under the same strong tracking policy.

\subsection{BABEL Long-Term Evaluation}

BABEL provides frame-level action annotations over long motion sequences. We use these annotations to construct streaming text-conditioned rollouts. Given a BABEL sequence with annotated segments, DAJI generates the corresponding motion segments sequentially. Each segment is conditioned on its text label and generated for the same duration as the annotation. The generated segment is appended to the latent history before generating the next segment, so later segments are conditioned on previously generated latents rather than generated independently and stitched together afterwards.

We evaluate each BABEL rollout from two perspectives. Subsequence metrics are computed on individual text-labeled motion segments and measure whether each generated segment matches its corresponding instruction. Transition metrics are computed on short clips centered around adjacent text-switching boundaries and measure whether the generated motion remains smooth and natural when the instruction changes. For a boundary at frame $b$, we extract a transition clip from $b-15$ to $b+15$, containing 15 frames before and 15 frames after the boundary.

\subsection{Transition Metrics}

Transition FID and Diversity are computed using the evaluator motion encoder, but only on boundary-centered transition clips. Transition FID measures whether generated transition clips follow the distribution of real transition clips. Transition Diversity measures the spread of generated transition clips in motion-feature space.

For transition smoothness, jerk is computed on the joint-position channels of the raw 38D robot representation, rather than on the full feature vector. Given a joint-position sequence $\mathbf{x}_t$, we compute
\begin{equation}
\mathbf{j}_t
=
\mathbf{x}_{t+3}
-
3\mathbf{x}_{t+2}
+
3\mathbf{x}_{t+1}
-
\mathbf{x}_t .
\end{equation}
Peak Jerk (PJ) is the maximum jerk magnitude within the transition clip:
\begin{equation}
\mathrm{PJ}
=
\max_t
\|\mathbf{j}_t\|_2 .
\end{equation}
Mean Jerk (MJ) measures the average jerk magnitude over the transition clip:
\begin{equation}
\mathrm{MJ}
=
\frac{1}{T_{\mathbf{x}}-3}
\sum_{t=1}^{T_{\mathbf{x}}-3}
\|\mathbf{j}_t\|_2 .
\end{equation}
PJ captures the largest abrupt motion change around the text-switching boundary, while MJ measures the overall smoothness of the transition.

\subsection{Rollout Success}

For HumanML3D-style evaluation, a generated motion is considered successful if it can be exported and validated in the robot-motion pipeline without triggering termination conditions. We use the same termination criteria as in controller evaluation, including body-height termination and gravity-direction termination. Success rate is computed as
\begin{equation}
\mathrm{Succ}
=
\frac{N_{\mathrm{success}}}
{N_{\mathrm{total}}}.
\end{equation}
Failed rollouts are counted in the denominator. Text-motion metrics are computed on successfully exported motions only, and success rate is reported separately to make invalid generations explicit.

\section{Implementation Details}
\label{sec:implementation_details}

\subsection{Key Dimensions and Training Constants}
\label{sec:key_dimensions}

Table~\ref{tab:key_dimensions} summarizes the key dimensions and temporal constants used by DAJI. Symbols are introduced in the main text when they first appear.

\begin{table}[h]
\centering
\small
\caption{Key dimensions and generation constants.}
\label{tab:key_dimensions}
\setlength{\tabcolsep}{8pt}
\begin{tabular}{lc}
\toprule
\textbf{Quantity} & \textbf{Value} \\
\midrule
Joint-intent latent dim $d_z$ & 64 \\
DAJI-Act control context dim $d_c$ & 192 \\
Full non-privileged training/distillation observation dim $d_{\mathrm{train}}$ & 1590 \\
Deployment proprioceptive/state observation dim $d_{\mathrm{prop}}$ & 812 \\
Training-only reference/motion observation dim $d_{\mathrm{ref}}$ & 778 \\
Raw privileged observation dim $d_{\mathrm{priv}}$ & 4558 \\
Critic-only privileged dim $d_{\mathrm{priv\_critic}}$ & 3 \\
Text hidden dim $d_{\mathrm{vlm}}$ & 2560 \\
Action dim $d_{\mathrm{act}}$ & 29 \\
\midrule
Control frequency & 50 Hz \\
Simulation timestep $\Delta t$ & 0.02 s \\
Latent chunk length $H$ & 15 frames (0.3 s) \\
Latent-history chunks $K$ & 8 \\
Max latent-history frames $T_{\mathrm{obs}}$ & 120 ($K \times H$) \\
Self-conditioning depth $K_{\mathrm{sc}}$ & 8 chunks \\
Default free-generation chunks $N$ & 20 \\
Euler integration steps $M$ & 4 \\
DDPM diffusion steps $T$ & 50 \\
\bottomrule
\end{tabular}
\end{table}

\FloatBarrier

\subsection{Architecture and Training Overview}
\label{sec:arch}

\subsubsection{Controller}

\textbf{Teacher.} The privileged encoder is an MLP $[512] \to 256$. The actor concatenates the non-privileged training observation with the privileged feature and passes through MLP $[1024, 512, 512]$, outputting the mean of an independent Gaussian with learned diagonal covariance over $d_{\mathrm{act}} = 29$ joint-position actions. The critic uses the same MLP architecture.

\textbf{DAJI-Act.} $E_{\mathrm{prop}}$ is a 2-layer MLP $[128, 128]$ with ELU activations, mapping the 812-dimensional deployment proprioceptive/state observation to a 128-dimensional feature. $E_{\mathrm{ref}}$ is a 2-layer MLP $[128, 128]$ with ELU activations, mapping the 778-dimensional training-only reference/motion observation to intent-distribution parameters $\boldsymbol{\mu}, \log\boldsymbol{\sigma}^2 \in \mathbb{R}^{64}$. The diffusion denoiser $\mathcal{D}_\phi$ follows a compact DDPM-style parameterization with deterministic DDIM sampling at inference~\cite{ho2020ddpm,song2021ddim}. It is implemented as a 4-layer MLP (width 256) with adaptive layer normalization (AdaLN): each residual block applies LayerNorm modulated by the time-condition embedding, followed by Linear(256)$\to$SiLU$\to$Linear(256) with a learned gate. The condition $\mathbf{c}_t \in \mathbb{R}^{192}$ is projected to 256d and summed with the time embedding. Xavier uniform initialization; final AdaLN modulation and output projection are zero-initialized. At deployment, DAJI-Act conditions only on $E_{\mathrm{prop}}(\mathbf{o}_t^{\mathrm{prop}})$ and the generated latent $\mathbf{z}_t$; the training-only reference observation is not used.

\subsubsection{DAJI-Flow}

\textbf{VLM.} Qwen3-VL-4B-Instruct~\cite{Bai2025Qwen3VLTR}, frozen. $d_{\mathrm{vlm}} = 2560$. Runs in bfloat16.

\textbf{Observation encoder $E_{\mathrm{obs}}$.} MLP: $d_z \to 768 \to d_{\mathrm{vlm}}$ with GELU activation, plus learned positional embeddings (max 120 positions), followed by LayerNorm. Truncates input to most recent $T_{\mathrm{obs}}$ frames.

\textbf{Flow-matching head.} DAJI-Flow predicts joint-intent latent chunks rather than low-level joint actions. We use a DiT-B flow-matching head~\cite{lipman2023flow,peebles2023dit} with 16 transformer blocks, 12 attention heads, token width 768, output dimension 1024, cross-attention to $d_{\mathrm{vlm}}$ conditioning tokens, AdaLayerNorm timestep modulation, and dropout 0.2. The noisy latent trajectory is embedded with sinusoidal flow-timestep encoding and learned latent-position embeddings. We prepend 32 learnable query tokens for internal aggregation, but only the $H=15$ latent-frame tokens are decoded as the output chunk by an MLP $1024 \to 1024 \to d_z$.

\subsubsection{Training Stages}

Table~\ref{tab:training_stages} summarizes which modules are trained or frozen in each of the three stages.

\begin{table}[h]
\centering
\small
\caption{Training stage breakdown. $\checkmark$ = trained; $\times$ = frozen; $-$ = not used.}
\label{tab:training_stages}
\setlength{\tabcolsep}{6pt}
\begin{tabular}{lccc}
\toprule
\textbf{Module} & \textbf{1. Teacher} & \textbf{2. Distill} & \textbf{3. Generator} \\
\midrule
Teacher (PPO actor + critic) & $\checkmark$ & $\times$ & $-$ \\
Intent encoder + DAJI-Act diffusion head & $-$ & $\checkmark$ & $-$ \\
Qwen VLM & $-$ & $-$ & $\times$ \\
LatentObsEncoder + DiT head & $-$ & $-$ & $\checkmark$ \\
\midrule
Data source & Simulation & Simulation & Latent dataset \\
Supervision & PPO reward & Teacher actions & Velocity field \\
\bottomrule
\end{tabular}
\end{table}

\FloatBarrier

\subsection{Observation, Privileged Inputs, and Rewards}
\label{sec:obs_rewards}

\subsubsection{Observation and Privileged Inputs}
\label{sec:obs}

Table~\ref{tab:obs_prop} reports the deployment proprioceptive/state observation used by DAJI-Act at test time. During teacher/distillation training, this observation is paired with the training-only reference/motion block in Table~\ref{tab:obs_ref} to form the 1590-dimensional non-privileged training observation.

\begin{table}[h]
\centering
\begin{minipage}{0.78\linewidth}
\centering
\small
\caption{Deployment proprioceptive/state observation used by DAJI-Act. For compactness, the 9-frame history uses offsets $\{0,1,2,3,4,8,12,16,20\}$.}
\label{tab:obs_prop}
\setlength{\tabcolsep}{6pt}
\renewcommand{\arraystretch}{1.10}
\begin{tabular*}{\linewidth}{l@{\extracolsep{\fill}}lc}
\toprule
\textbf{Observation block} & \textbf{Temporal form} & \textbf{Dim.} \\
\midrule
Root angular velocity & $3$D body-frame vector over 9 offsets & 27 \\
Projected gravity & $3$D body-frame vector over 9 offsets & 27 \\
Joint positions & $29$D joint vector over 9 offsets & 261 \\
Joint velocities & $29$D joint-velocity vector over 9 offsets & 261 \\
Previous actions & $29$D action vector over 8 history steps & 232 \\
Boot indicator & scalar flag & 1 \\
Compliance flag & raw, threshold-scaled, and stiffness-scaled flags & 3 \\
\midrule
\textbf{Total} & -- & \textbf{812} \\
\bottomrule
\end{tabular*}
\end{minipage}
\end{table}

\begin{table}[h]
\centering
\begin{minipage}{0.78\linewidth}
\centering
\small
\caption{Training-only reference/motion observation used by the intent encoder and privileged teacher during distillation.}
\label{tab:obs_ref}
\setlength{\tabcolsep}{6pt}
\renewcommand{\arraystretch}{1.10}
\begin{tabular*}{\linewidth}{l@{\extracolsep{\fill}}lc}
\toprule
\textbf{Observation block} & \textbf{Temporal form} & \textbf{Dim.} \\
\midrule
Command/reference state & root pose-related features & 96 \\
Target joint positions & future reference features & 638 \\
Target root height & 11 future steps & 11 \\
Target projected gravity & $3$ axes $\times$ 11 future steps & 33 \\
\midrule
\textbf{Total} & -- & \textbf{778} \\
\bottomrule
\end{tabular*}
\end{minipage}
\end{table}

The full non-privileged observation used during teacher/distillation training has dimension 1590, combining the 812-dimensional proprioceptive/state block with the 778-dimensional reference-motion block. At deployment, DAJI-Act does not consume this full 1590-dimensional input. Instead, it conditions on the 812-dimensional proprioceptive/state observation together with the generated joint-intent latent. During training, the privileged teacher and critic additionally use simulation-only privileged inputs, which are not available at deployment.

In our implementation, the teacher actor receives the full non-privileged training observation together with a 256-dimensional privileged feature encoded from the raw privileged observation. The critic receives the concatenation of the full non-privileged training observation, the raw privileged observation, and a critic-only error signal. The raw privileged observation has dimension 4558, and the critic-only privileged signal has dimension 3.

The non-privileged training observation contains deployable and task-related terms, including boot indicators, command/reference observations, target root height, target joint positions, target projected gravity, root angular-velocity history, projected-gravity history, joint-position history, joint-velocity history, and previous actions. Only the proprioceptive/state subset is available to deployed DAJI-Act.

The raw privileged observation provides simulation-only information, including future target root positions and velocities in the body frame, target-current root orientation difference, body-height signals, foot contact forces, true root linear velocity, longer proprioceptive histories, current keypoint positions and velocities, target-current keypoint differences, applied actions, and applied torques. The critic-only privileged signal contains cumulative tracking errors over root, orientation, and keypoint terms.

These privileged inputs are used to train the privileged teacher encoder and critic, but they are not available to the deployed DAJI-Act actor.

\subsubsection{Rewards and Domain Randomization}
\label{sec:rewards}

\begin{table}[h]
\centering
\begin{minipage}{0.6\linewidth}
\centering
\small
\caption{Tracking reward terms. All tracking rewards use an exponential kernel $\exp(-\|e\| / \sigma)$.}
\label{tab:rewards_tracking}
\setlength{\tabcolsep}{6pt}
\renewcommand{\arraystretch}{1.08}
\begin{tabular*}{\linewidth}{l@{\extracolsep{\fill}}cc}
\toprule
\textbf{Reward term} & \textbf{Weight} & $\boldsymbol{\sigma}$ \\
\midrule
Root position & 0.5 & 0.3 \\
Root rotation (axis-angle) & 0.5 & 0.4 \\
Root linear velocity & 1.0 & 1.0 \\
Root angular velocity & 1.0 & 3.0 \\
Keypoint position (13 bodies) & 1.0 & 0.3 \\
Keypoint velocity & 1.0 & 1.0 \\
Keypoint rotation (6D) & 1.0 & 0.4 \\
Keypoint angular velocity & 1.0 & 3.0 \\
Lower-body keypoints & 0.5 & 0.2 \\
Upper-body keypoints & 0.5 & 0.3 \\
Joint position (29 DoF) & 1.0 & 0.5 \\
Joint velocity & 0.5 & 3.0 \\
\bottomrule
\end{tabular*}
\end{minipage}
\end{table}

\begin{table}[h]
\centering
\begin{minipage}{0.6\linewidth}
\centering
\small
\caption{Regularization reward terms.}
\label{tab:rewards_regularization}
\setlength{\tabcolsep}{6pt}
\renewcommand{\arraystretch}{1.08}
\begin{tabular*}{\linewidth}{l@{\extracolsep{\fill}}c}
\toprule
\textbf{Reward term} & \textbf{Weight} \\
\midrule
Survival bonus & 3.0 \\
Joint velocity L2 & $5{\times}10^{-4}$ \\
Action rate L2 & 0.01 \\
Feet air time (ref-aligned) & 5.0 \\
Feet air time (dense) & 1.0 \\
Joint position limits (soft) & 1.0 \\
Joint torque limits (soft) & 0.01 \\
Joint acceleration L2 (teacher only) & $4{\times}10^{-8}$ \\
\bottomrule
\end{tabular*}
\end{minipage}
\end{table}

\paragraph{Domain randomization.}
\label{sec:domain_rand}

\begin{table}[h]
\centering
\begin{minipage}{0.6\linewidth}
\centering
\small
\caption{Domain randomization ranges grouped by subsystem.}
\label{tab:randomization}
\setlength{\tabcolsep}{6pt}
\renewcommand{\arraystretch}{1.08}
\begin{tabular*}{\linewidth}{l@{\extracolsep{\fill}}l}
\toprule
\textbf{Randomization} & \textbf{Range} \\
\midrule
\multicolumn{2}{l}{\textit{Contact and solver}} \\
\addlinespace[2pt]
CoM perturbation (pelvis, torso) & $\pm 0.03~\mathrm{m}$ \\
Ankle friction & $0.3$--$1.2$ \\
Solver time constant & $0.015$--$0.03$ \\
Solver damping ratio & $0.5$--$2.0$ \\

\addlinespace[4pt]
\multicolumn{2}{l}{\textit{Motor and joint}} \\
\addlinespace[2pt]
Motor stiffness (arms) & $0.9$--$1.1$ \\
Motor stiffness (legs) & $0.5$--$2.0$ \\
Motor damping (arms) & $0.9$--$1.1$ \\
Motor damping (legs) & $0.5$--$2.0$ \\
Motor armature & $0.75$--$1.25$ \\
Joint offset noise & $\pm 0.01~\mathrm{rad}$ \\
Joint pos. bias (hip, knee, ankle) & $\sigma = 0.1~\mathrm{rad}$ \\

\addlinespace[4pt]
\multicolumn{2}{l}{\textit{Root perturbation}} \\
\addlinespace[2pt]
Root drift velocity (xy) & $\pm 0.25~\mathrm{m/s}$ \\
Root drift velocity (z) & $\pm 0.05~\mathrm{m/s}$ \\
Root height offset & $\pm 0.03~\mathrm{m}$ \\
Root velocity push (xy) & $\pm 0.5~\mathrm{m/s}$ \\
Root velocity push (z) & $\pm 0.2~\mathrm{m/s}$ \\
Root velocity push (roll, pitch) & $\pm 0.52~\mathrm{rad}$ \\
Root velocity push (yaw) & $\pm 0.78~\mathrm{rad}$ \\
Push interval & $4.0$--$6.0~\mathrm{s}$ \\
Gravity perturbation & $\sigma = 0.1~\mathrm{m/s^2}$ \\
\bottomrule
\end{tabular*}
\end{minipage}
\end{table}

\subsection{Training Details}
\label{sec:training_details}

\subsubsection{Training Hyperparameters}
\label{sec:hparams}

Tables~\ref{tab:controller_hp} and~\ref{tab:generator_hp} summarize the hyperparameters used for controller training, distillation, and DAJI-Flow generation. Unless otherwise specified, these settings are used for all main experiments and ablations.

\begin{table}[h]
\centering
\begin{minipage}{0.78\linewidth}
\centering
\small
\caption{Controller training hyperparameters.}
\label{tab:controller_hp}
\setlength{\tabcolsep}{6pt}
\renewcommand{\arraystretch}{1.08}
\begin{tabular*}{\linewidth}{l@{\extracolsep{\fill}}cc}
\toprule
\textbf{Parameter} & \textbf{Teacher (PPO)} & \textbf{Distill} \\
\midrule
Discount $\gamma$ & 0.99 & \\
GAE $\lambda$ & 0.95 & \\
PPO clip $\epsilon$ & 0.2 & \\
Entropy coefficient & $0.01 \to 0.0025$ (linear) & 0 \\
Learning rate & $5{\times}10^{-4}$ & $1{\times}10^{-4}$ \\
PPO epochs & 5 & \\
Optimization epochs per buffer & & 5 \\
Minibatches & 8 & 8 \\
Gradient clipping & 1.0 & 1.0 \\
Total env frames & $8{\times}10^9$ & $\sim\!8{\times}10^9$ \\
\midrule
KL weight $\beta$ & & $1{\times}10^{-4}$ \\
Free bits & & 0.0 \\
Inference mode & & mean (deterministic) \\
DDPM steps $T$ & & 50 \\
Noise schedule & & Cosine \\
DDIM sampling steps & & 2 \\
DDIM $\eta$ & & 0 (deterministic) \\
\midrule
Symmetry loss weight (loc) & 0.2 & \\
Symmetry loss weight (scale) & 10.0 & \\
\bottomrule
\end{tabular*}
\end{minipage}
\end{table}

\begin{table}[h]
\centering
\begin{minipage}{0.78\linewidth}
\centering
\small
\caption{Generator hyperparameters.}
\label{tab:generator_hp}
\setlength{\tabcolsep}{6pt}
\renewcommand{\arraystretch}{1.08}
\begin{tabular*}{\linewidth}{l@{\extracolsep{\fill}}c}
\toprule
\textbf{Parameter} & \textbf{Value} \\
\midrule
$K$ (latent-history chunks) & 8 \\
$H$ (latent chunk length) & 15 frames (current + 14 future) \\
$T_{\mathrm{obs}}$ (max observation frames) & 120 ($K \times H$) \\
Self-conditioning depth & 8 chunks \\
$M$ (Euler steps) & 4 \\
$N$ (default chunks per free rollout) & 20 \\
Flow time distribution & $\mathrm{Beta}(1.5, 1.0)$, rescaled to $[0, 0.999]$ \\
Timestep buckets & 1000 \\
\midrule
Obs encoder LR & $1{\times}10^{-4}$ \\
DiT generator LR & $1{\times}10^{-4}$ \\
LR schedule & Cosine, min\_lr $5{\times}10^{-7}$ \\
Warmup steps & 5000 \\
Gradient clipping & 1.0 \\
Mixed precision & bf16 (VLM), fp32 (DiT) \\
Max training steps & 200,000 \\
\bottomrule
\end{tabular*}
\end{minipage}
\end{table}

\subsubsection{Scheduled Self-Conditioned Training}
\label{sec:self_conditioning}

DAJI-Flow is trained with scheduled self-conditioning to reduce the train-test mismatch of autoregressive latent generation. During standard teacher-forced training, the generator observes ground-truth latent histories. During deployment, however, it must condition on its own previously generated latent chunks. Scheduled self-conditioning exposes the generator to such predicted-history inputs during training.

Given a ground-truth latent trajectory $\mathbf{Z}_{1:T}$, we sample a starting index $t$ and take the recent ground-truth latent history as the initial history buffer:
\begin{equation}
\mathcal{H}^{(0)}
=
\mathbf{Z}_{t-T_{\mathrm{obs}}:t}.
\end{equation}
The generator is then unrolled for $K_{\mathrm{sc}}$ chunks. In our default setting, $K_{\mathrm{sc}}=8$ and each chunk contains $H=15$ latent frames, corresponding to an 8-chunk, 120-frame unrolled training horizon.

For the $k$-th chunk, the ground-truth target is
\begin{equation}
\mathbf{X}^{(k)}_0
=
\mathbf{Z}_{t+(k-1)H:t+kH}.
\end{equation}
DAJI-Flow predicts this chunk using the current history buffer $\mathcal{H}^{(k-1)}$ and the language condition. The first chunk is conditioned on ground-truth history. For subsequent chunks, the history buffer is augmented with previously generated latent chunks:
\begin{equation}
\mathcal{H}^{(k)}
=
\mathrm{Update}
\left(
\mathcal{H}^{(k-1)},
\mathrm{sg}\bigl(\hat{\mathbf{X}}^{(k)}_0\bigr)
\right),
\end{equation}
where $\mathrm{sg}(\cdot)$ denotes stop-gradient. Thus, generated chunks are used as predicted conditioning context, but gradients are not back-propagated through earlier sampling steps or ODE integrations.

The self-conditioned training loss combines the primary one-step flow-matching loss with the average loss over later self-conditioned chunks:
\begin{equation}
\mathcal{L}_{\mathrm{SC}}
=
\mathcal{L}_{\mathrm{Flow}}^{(1)}
+
\lambda_{\mathrm{sc}}
\frac{1}{K_{\mathrm{sc}}-1}
\sum_{k=2}^{K_{\mathrm{sc}}}
\mathcal{L}_{\mathrm{Flow}}^{(k)}.
\end{equation}
Here, $\mathcal{L}_{\mathrm{Flow}}^{(1)}$ is the standard teacher-forced flow-matching loss for the first chunk, while $\mathcal{L}_{\mathrm{Flow}}^{(k)}$ for $k \ge 2$ is computed using histories augmented with previously generated chunks. We use $\lambda_{\mathrm{sc}}=0.25$.

We apply this objective according to a training curriculum. Let $u$ denote the current training step. Self-conditioning is disabled for the first 100k training steps:
\begin{equation}
p_{\mathrm{sc}}(u)=0,
\qquad u < 100{,}000.
\end{equation}
It is then linearly increased to 1 over the next 40k steps:
\begin{equation}
p_{\mathrm{sc}}(u)
=
\frac{u-100{,}000}{40{,}000},
\qquad
100{,}000 \le u < 140{,}000.
\end{equation}
Afterwards, self-conditioning is always enabled:
\begin{equation}
p_{\mathrm{sc}}(u)=1,
\qquad u \ge 140{,}000.
\end{equation}
At each training step, the self-conditioned objective is used with probability $p_{\mathrm{sc}}(u)$; otherwise, the generator is trained with the standard single-chunk teacher-forced flow-matching objective.

This training strategy does not change the generator architecture. It only changes the distribution of histories observed during training, making them closer to the predicted histories encountered during autoregressive deployment.

\FloatBarrier

\section{Additional Experiments and Analysis}
\label{sec:additional_experiments}

\subsection{Representation-Level Latent Diagnostics}
\label{sec:latent_diagnostics}

This section supplements the main-paper latent temporal design ablation by defining the representation-level diagnostics used in Table~\ref{tab:latent_design_probe}. We use these diagnostics to evaluate whether the learned joint-intent latent contains future-relevant information, evolves coherently over time, and remains executable in closed-loop control.

\subsubsection{Future-Prediction Probe}

Probe@40 evaluates whether the current latent contains information about future robot motion 40 frames later. Given the current 64-dimensional joint-intent latent $\mathbf{z}_t$ and robot joint positions $\mathbf{q}_t$, we construct a future joint-motion target from the residual
\begin{equation}
\Delta \mathbf{q}_{t,40}
=
\mathbf{q}_{t+40}
-
\mathbf{q}_{t}.
\end{equation}
To obtain a compact target, we project these joint residuals onto their leading PCA components. We then train a linear ridge probe to predict this future-motion target from $\mathbf{z}_t$. The probe is trained on the training split and evaluated on held-out motion sequences. Probe@40 reports the Pearson correlation between the predicted and ground-truth future-motion targets on held-out sequences. Higher values indicate that the current latent contains more predictive information about upcoming motion.

\subsubsection{Latent Temporal Correlation}

Corr.@40 measures the temporal coherence of the latent trajectory. We compute the cosine similarity between latents separated by 40 frames:
\begin{equation}
\mathrm{Corr.@40}
=
\mathbb{E}_{t}
\left[
\frac{
\mathbf{z}_t^\top \mathbf{z}_{t+40}
}{
\|\mathbf{z}_t\|_2
\|\mathbf{z}_{t+40}\|_2
}
\right].
\end{equation}
The score is averaged over valid frames and motion sequences. Higher values indicate that the latent evolves more smoothly and consistently over time.

\subsubsection{Closed-Loop Success at Fixed Horizons}

Succ.@20s and Succ.@60s measure long-horizon closed-loop executability. For each latent temporal design, we evaluate the corresponding DAJI-Act controller in simulation for fixed horizons of 20 seconds and 60 seconds. A rollout is counted as successful if it reaches the evaluation horizon without triggering termination conditions such as body-height or gravity-direction failure. We report the percentage of successful rollouts.

\subsection{Hyperparameter Sensitivity}
\label{sec:hyperparam_sensitivity_appendix}

We analyze the sensitivity to self-conditioning depth and latent-history length on the HumanML3D-style benchmark. These results show that DAJI is not tied to a fragile hyperparameter choice, but works across a reasonably broad operating regime.

\begin{table}[h]
\centering
\caption{\textbf{Hyperparameter sensitivity on HumanML3D.}
Each history chunk contains 15 frames. In each block, we vary one factor while keeping the remaining settings fixed.}
\label{tab:hyperparam_sensitivity}
\vspace{0.3em}
\setlength{\tabcolsep}{4pt}
\renewcommand{\arraystretch}{1.08}
\begin{tabular}{lcccccc}
\toprule
\multicolumn{7}{c}{Self-Conditioning Depth} \\
\midrule
Depth & MM-D $\downarrow$ & R@1 $\uparrow$ & R@3 $\uparrow$ & FID $\downarrow$ & MM $\uparrow$ & Succ. (\%) $\uparrow$ \\
\midrule
4 chunks & 1.181 & 0.398 & 0.618 & 0.176 & 0.516 & 93.34 \\
6 chunks & 1.115 & 0.483 & 0.754 & 0.148 & 0.689 & 93.47 \\
8 chunks & \textbf{1.093} & \textbf{0.549}  & \textbf{0.796} & \textbf{0.147} & \textbf{0.847} & \textbf{94.42} \\
\midrule
\multicolumn{7}{c}{Latent History Length} \\
\midrule
History & MM-D $\downarrow$ & R@1 $\uparrow$ & R@3 $\uparrow$ & FID $\downarrow$ & MM $\uparrow$ & Succ. (\%) $\uparrow$ \\
\midrule
K=4 (60f)  & 1.132 & 0.480 & 0.743 & 0.177 & 0.792 & 93.38 \\
K=8 (120f) & \textbf{1.093} & \textbf{0.549}  & \textbf{0.796} & \textbf{0.147} & \textbf{0.847} & \textbf{94.42} \\
K=12 (180f) & 1.129 & 0.478 & 0.731 & 0.170 & 0.587 & 93.70 \\
\bottomrule
\end{tabular}
\end{table}

Table~\ref{tab:hyperparam_sensitivity} shows that the main trends are stable. Increasing self-conditioning depth improves semantic alignment and rollout success, which is consistent with reducing exposure bias over longer rollout horizons. For latent-history length, $K=8$ gives the strongest overall tradeoff across alignment, diversity, and success, while both shorter and longer histories underperform on at least part of the metric set. We therefore use eight self-conditioned chunks during training and $K=8$ latent-history chunks at inference.

\subsection{Qualitative Generation Results}
\label{sec:qualitative_generation_results}

\begin{figure}[p]
\centering
\newcommand{\twocolrow}[4]{%
  \begin{minipage}[t]{0.48\linewidth}
    \centering
    \includegraphics[width=\linewidth]{#1}\\[6pt]
    {\footnotesize\textit{``#2''}}
  \end{minipage}\hfill
  \begin{minipage}[t]{0.48\linewidth}
    \centering
    \includegraphics[width=\linewidth]{#3}\\[6pt]
    {\footnotesize\textit{``#4''}}
  \end{minipage} \\[12pt]
}
\begin{minipage}[t]{0.48\linewidth}
  \centering \textbf{Sim (MuJoCo)}
\end{minipage}\hfill
\begin{minipage}[t]{0.48\linewidth}
  \centering \textbf{Sim (MuJoCo)}
\end{minipage} \\[10pt]

\twocolrow{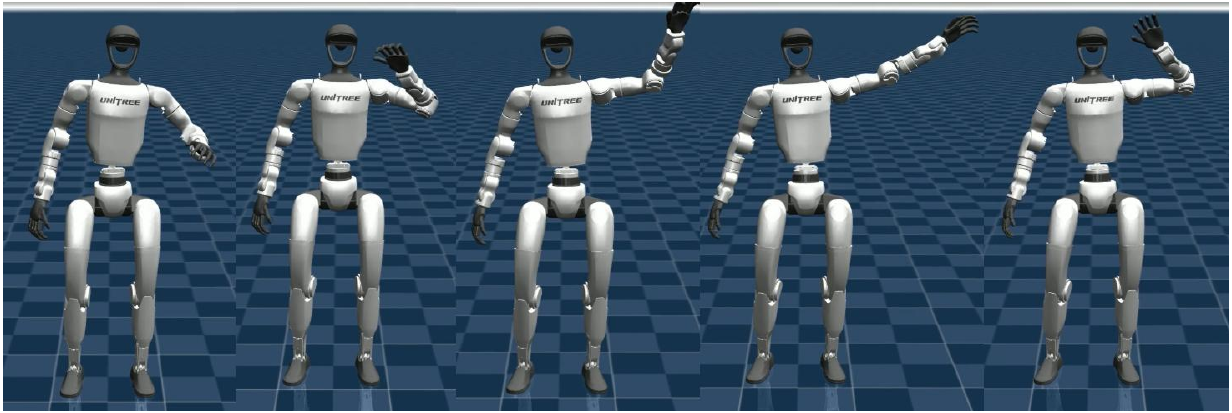}{A man is waving his left hand.}%
          {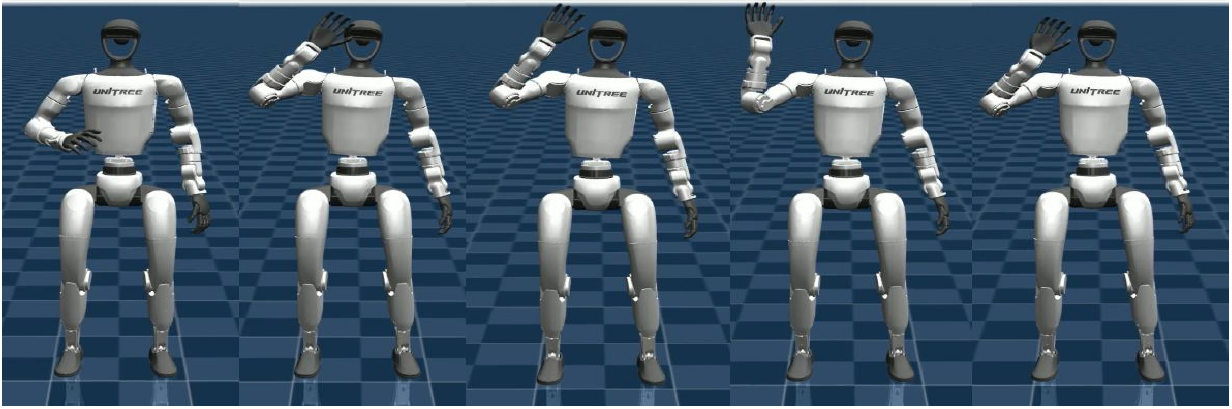}{A man is waving his right hand.}

\twocolrow{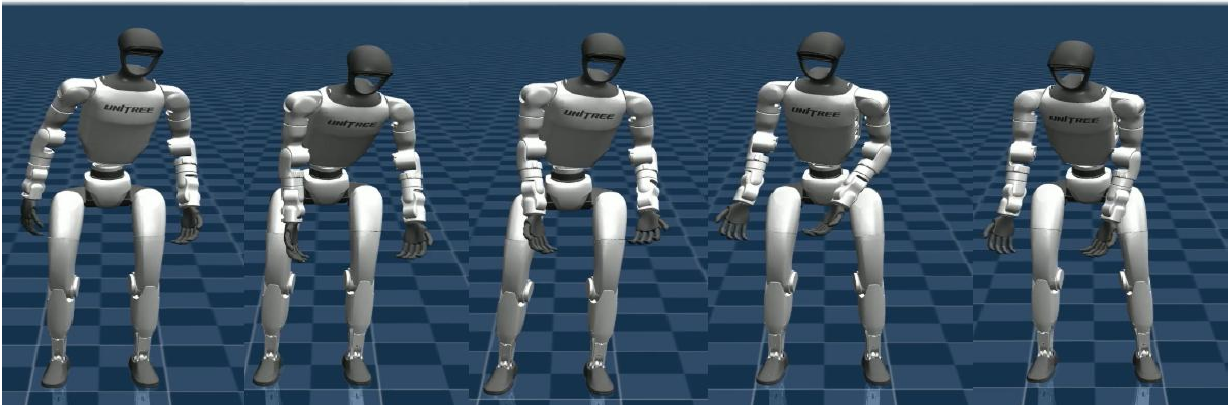}{A person reaches from the left side to the right side.}%
          {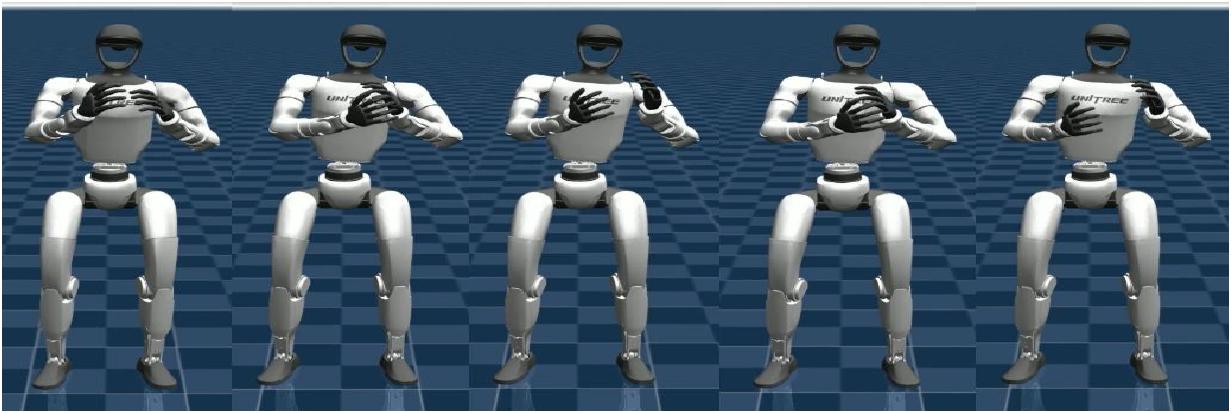}{A person claps his hands.}

\twocolrow{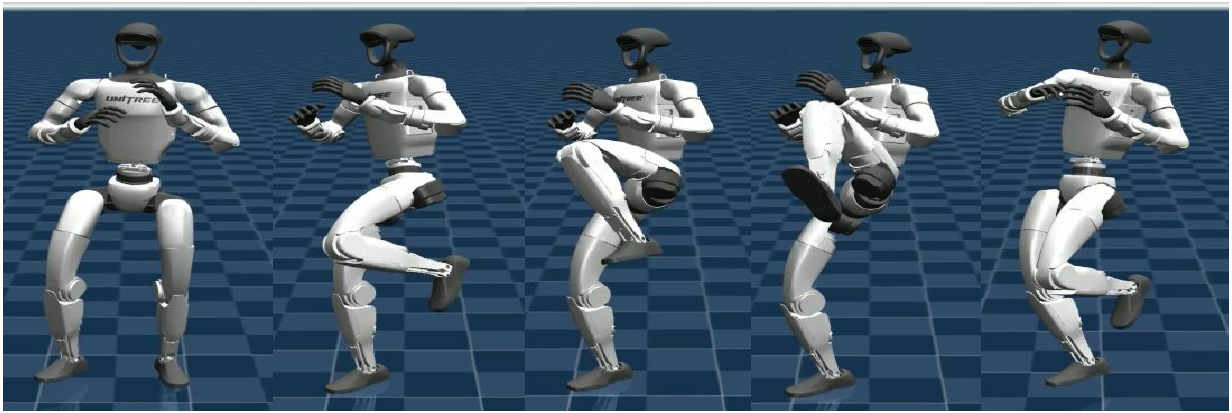}{A person flies kick.}%
          {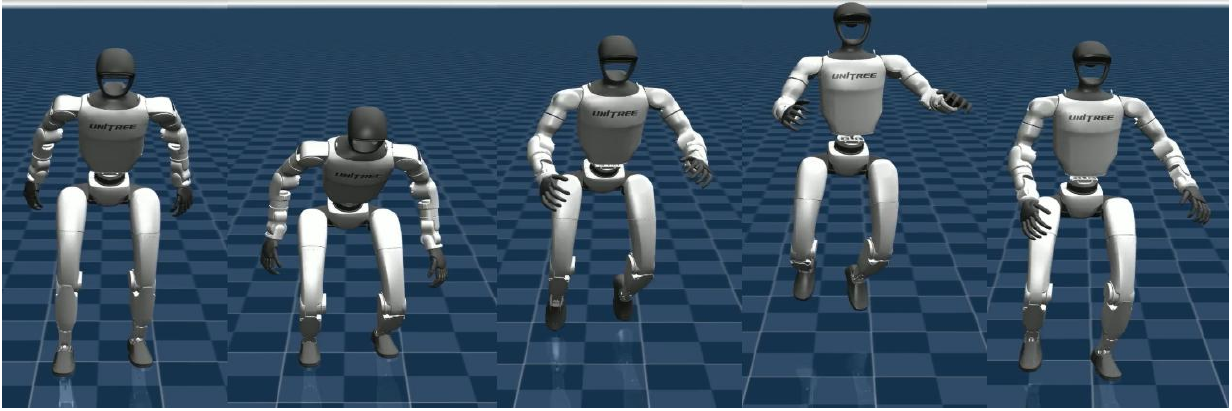}{A person is jumping forward.}

\twocolrow{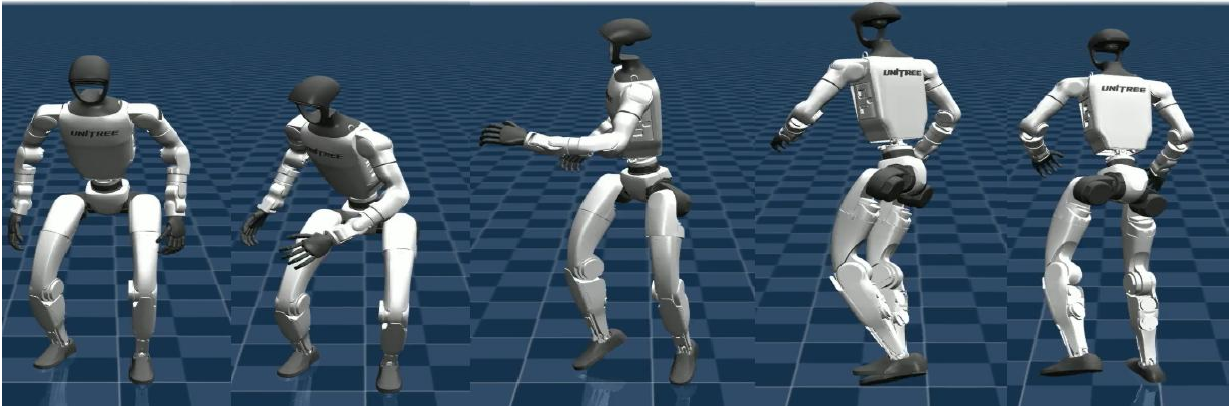}{A man jumps up in a tight twirl.}%
          {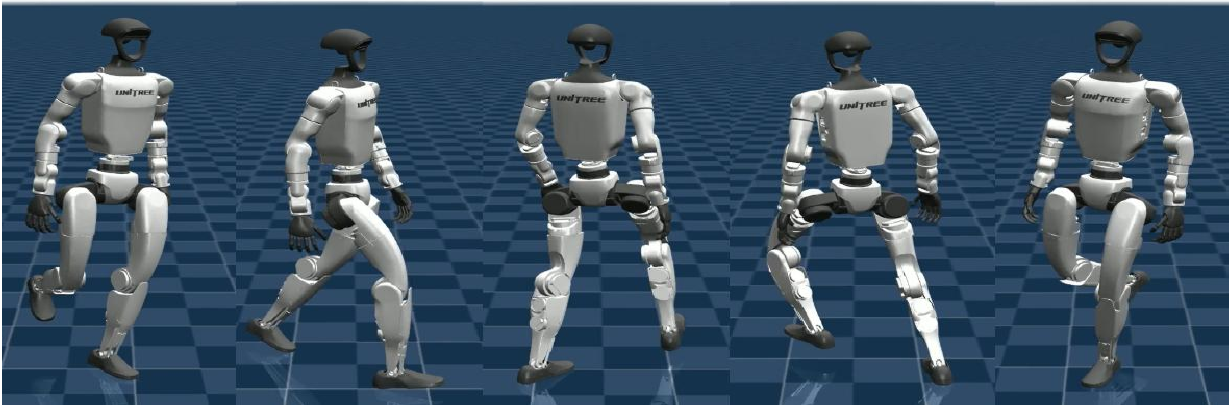}{A person walks in a circle.}

\twocolrow{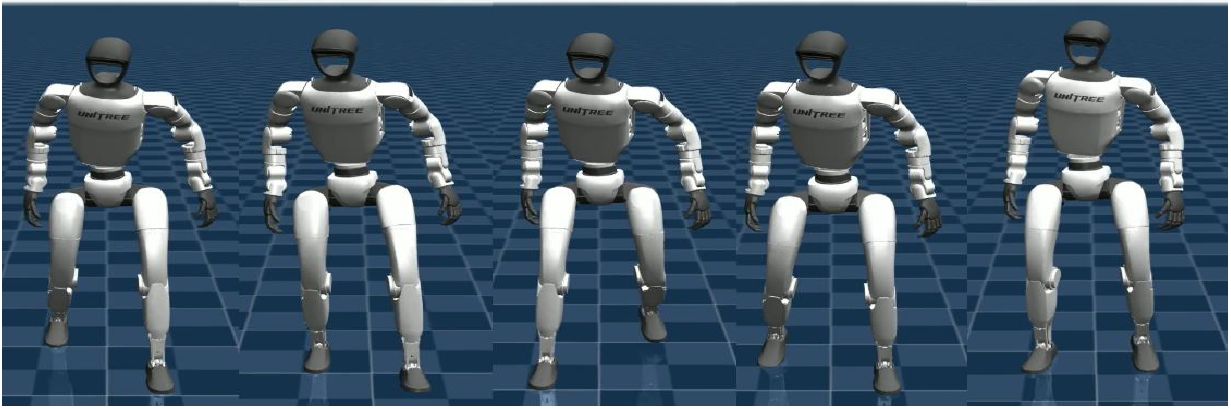}{A person is walking backward.}%
          {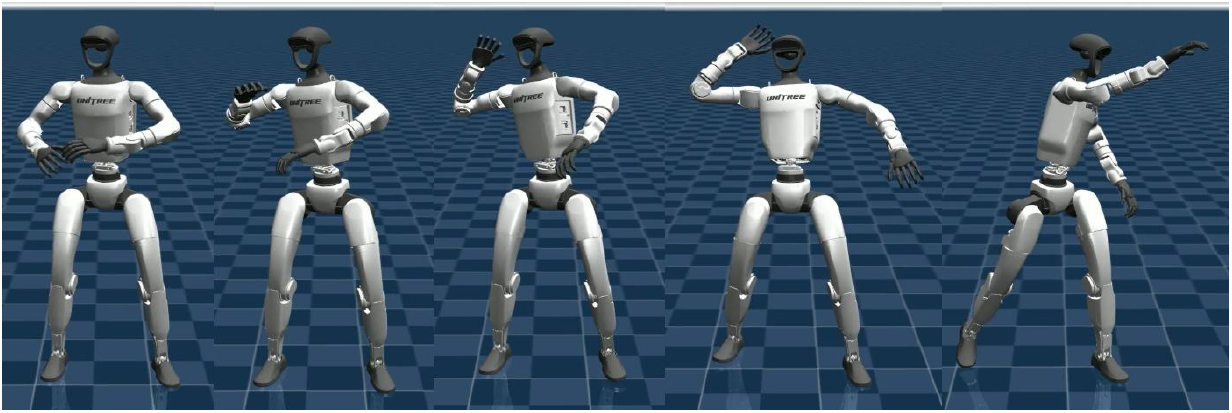}{The person is throwing a baseball.}

\vspace{2pt}
\caption{\textbf{Sim (MuJoCo) Validation Results:} Robust tracking performance on simple gestures to complex maneuvers. Any prompt phrase is interpreted only as a body-motion description; no object state or manipulation outcome is modeled or evaluated.}
\label{fig:sim_mujoco}
\end{figure}

Figure~\ref{fig:sim_mujoco} provides representative text-conditioned execution examples within the MuJoCo~\cite{todorov2012mujoco} simulation environment, illustrating robust tracking across a variety of dynamic and complex whole-body maneuvers.

\FloatBarrier

\section{Additional Information}
\label{sec:additional_information}

\subsection{Extended Formulas}
\label{sec:ext_formulas}

\subsubsection{DDIM Reverse Step}

The deterministic DDIM \cite{10.1145/3746027.3755441, li2026deltascoremattersspatial,li2026oraclenoisefastersemantic, li2026z2samplingzerocostzigzagtrajectories, chen2025polarisprojectionorthogonalsquaresrobust, ning2025dctdiffintriguingpropertiesimage, 10.1145/3746027.3758161} reverse step ($\eta = 0$) from $\tau$ to $\tau'$ given the predicted $\hat{\mathbf{x}}_0$:
\begin{equation}
\begin{aligned}
\hat{\boldsymbol{\epsilon}} &= \frac{\mathbf{x}_\tau - \sqrt{\bar{\alpha}_\tau}\, \hat{\mathbf{x}}_0}{\sqrt{1 - \bar{\alpha}_\tau}}, \\
\mathbf{x}_{\tau'} &= \sqrt{\bar{\alpha}_{\tau'}}\, \hat{\mathbf{x}}_0 + \sqrt{1 - \bar{\alpha}_{\tau'}}\, \hat{\boldsymbol{\epsilon}}.
\end{aligned}
\end{equation}

\subsubsection{Intent Bottleneck KL Regularization}

The KL regularization for the intent encoder $E_{\mathrm{ref}}$:
\begin{equation}
\mathcal{L}_{\mathrm{KL}} = -\frac{1}{2} \sum_{j=1}^{d_z} \bigl(1 + \log \sigma_{t,j}^2 - \mu_{t,j}^2 - \sigma_{t,j}^2 \bigr).
\end{equation}
optionally clamped at a free-bits threshold to prevent posterior collapse.

\subsubsection{Flow-Matching Time Sampling}

The flow time $s$ is sampled as $s = 0.999 \cdot u$ with $u \sim \mathrm{Beta}(1.5, 1.0)$, then treated as $s \in [0,1]$ in the main text for notational simplicity. This distribution biases samples toward larger flow times, placing more emphasis on less noisy states closer to the data endpoint.

\subsubsection{Cosine Noise Schedule (DDPM)}

The forward-process variance $\bar{\alpha}_\tau$ with $T = 50$:
\begin{equation}
\bar{\alpha}_\tau = \frac{f(\tau)}{f(0)}, \qquad f(\tau) = \cos^2\!\Bigl(\frac{\tau/T + 0.008}{1.008} \cdot \frac{\pi}{2}\Bigr).
\end{equation}

\FloatBarrier

\subsection{Existing Assets and Terms of Use}
\label{sec:asset_licenses}

Table~\ref{tab:asset_licenses} lists the main external assets used in this work together with their stated licenses or usage terms. For datasets with access restrictions or downstream dependencies, we follow the terms of the official distribution source rather than redistributing the underlying files.

\begin{table}[h]
\centering
\small
\caption{Main external assets used in this work.}
\label{tab:asset_licenses}
\setlength{\tabcolsep}{4pt}
\renewcommand{\arraystretch}{1.12}
\begin{tabularx}{\linewidth}{p{2.2cm}p{2.7cm}Y p{3.0cm}}
\toprule
\textbf{Asset} & \textbf{Role} & \textbf{Source / version} & \textbf{License / terms} \\
\midrule
HumanML3D 
& Single-instruction text-motion benchmark 
& Official HumanML3D release~\cite{guo2022generating} 
& MIT License \\

BABEL 
& Long-horizon streaming action annotations 
& Official BABEL release~\cite{punnakkal2021babel} 
& BABEL dataset license for non-commercial scientific research \\

Qwen3-VL-4B-Instruct
& Frozen language encoder 
& Qwen3-VL-4B-Instruct checkpoint; technical report~\cite{Bai2025Qwen3VLTR} 
& Apache License 2.0 \\

MuJoCo 
& Physics simulation environment 
& MuJoCo simulator~\cite{todorov2012mujoco} 
& Apache License 2.0 \\
\bottomrule
\end{tabularx}
\end{table}

\subsection{Broader Impact}
\label{sec:broader_impact}

Language-conditioned humanoid control could benefit assistive robotics, animation, embodied AI prototyping, and human--robot interaction by making whole-body behavior easier to specify. However, a system that maps language to humanoid motion also creates clear risks if deployed in safety-critical or unstructured physical environments: ambiguous or adversarial instructions can trigger unintended behavior, generated motions may be dynamically executable in simulation yet unsafe on hardware, and the same capabilities could be misused for harmful physical tasks. Our large-scale quantitative experiments are limited to simulation and controlled robot-motion validation, while physical-hardware deployment is used for qualitative demonstration. Real-world deployment would require instruction filtering, hardware-level safety constraints, perception robustness, human supervision, and fail-safe monitoring that are outside the scope of this paper.

\subsection{Limitations}
\label{sec:limitations}

DAJI has several important limitations. First, although DAJI includes qualitative deployment on physical humanoid hardware, our large-scale quantitative evaluations are conducted in simulation and controlled robot-motion validation. Therefore, the reported quantitative results should be interpreted with the sim-to-real gap in mind, rather than as evidence of fully safe autonomous real-world deployment. Second, performance is constrained by the coverage and granularity of the paired motion-language datasets and by the frozen language encoder: rare actions, ambiguous instructions, long-horizon compositional commands, and object-centric phrases may be unreliable when they are underrepresented in training data or cannot be grounded in the available observations. Third, the current system is instruction-conditioned but not fully interactive. DAJI can generate streaming motions from language commands, but it does not yet support multi-turn user correction, clarification, or online adjustment of motion style during execution. Finally, our current evaluation does not report full multi-seed error bars for all experiments because full humanoid-control retraining is expensive, and some text-motion metrics are computed on successfully exported motions with rollout success reported separately. Improving dataset diversity, adding richer grounding and safety constraints, and incorporating interactive language feedback are important directions for future work.

\end{document}